\documentclass[lettersize,journal]{IEEEtran}

\IEEEoverridecommandlockouts                  

\usepackage{mathtools}
\usepackage{gensymb}
\usepackage[export]{adjustbox}
\usepackage{color}
\usepackage{graphicx}
\usepackage{booktabs}
\usepackage{epsfig} 
\usepackage{times} 
\usepackage{amsmath} 
\usepackage{amssymb}
\usepackage{amsthm}
\usepackage{leftidx}
\usepackage{stfloats}

\usepackage{siunitx}
\usepackage{nomencl}
\makenomenclature
\usepackage{hyperref}
\usepackage{csquotes}

\usepackage{enumitem}
\usepackage{tablefootnote}
\usepackage{ushort}
\usepackage{algpseudocode}
\usepackage[ruled,vlined]{algorithm2e}

\usepackage{stackengine}
\usepackage{multirow}  
\usepackage[caption=false,font=footnotesize]{subfig}

\title{ARCADE: Adaptive Robot Control with Online Changepoint-Aware Bayesian Dynamics Learning}

\author{
 Rishabh Dev Yadav$^{*1}$,
 Avirup Das$^{*1}$,
 Hongyu Song$^1$,
 Samuel Kaski$^{1,2}$,
 Wei Pan$^1$
 \thanks{$^*$ Authors contributed equally.}
 \thanks{$^1$ The authors are with the Department of Computer Science, The University of Manchester, United Kingdom. Contact:
       {\tt\footnotesize \{avirup.das, rishabh.yadav, hongyu.song-3\}@postgrad.manchester.ac.uk; \{samuel.kaski, wei.pan\}@manchester.ac.uk.}}
 \thanks{$^2$ The authors are with the Department of Computer Science, Aalto University, Finland.}
}

\begin{document}

\newcommand{\bmx}[1]{\begin{bmatrix}#1\end{bmatrix}}
\newcommand{\Bmx}[1]{\begin{bmatrix*}[r]#1\end{bmatrix*}}
\newcommand{\mx}[1]{\begin{matrix*}[r]#1\end{matrix*}}
\newcommand{\pmx}[1]{\begin{pmatrix}#1\end{pmatrix}}
\newcommand{\Pmx}[1]{\begin{pmatrix*}[r]#1\end{pmatrix*}}
\newcommand{\smx}[1]{\begin{Bmatrix}#1\end{Bmatrix}}
\newcommand{\vmx}[1]{\begin{vmatrix}#1\end{vmatrix}}
\newcommand{\Vmx}[1]{\begin{vmatrix*}[r]#1\end{vmatrix*}}
\newcommand{\norm}[1]{\left\lVert#1\right\rVert}
\newcommand{\abs}[1]{\left\lvert #1 \right\rvert}
\newcommand{\fbr}[1]{\left(#1\right)}
\newcommand{\tbr}[1]{\left[#1\right]}
\newcommand{\sbr}[1]{\left\{#1\right\}}
\newcommand{\ips}[1]{\left\langle#1\right\rangle}
\newcommand{\expectation}[2]{\mathop{\mathbb{E}}_{#2}\tbr{#1}}
\newcommand{\variance}[2]{\mathop{\mathbb{V}ar}_{#2}\tbr{#1}}
\newcommand{\Var}[1]{\operatorname{Var}\tbr{#1}}
\newcommand{\diag}[1]{\textrm{diag}\left(#1\right)}

\newtheorem{theorem}{Theorem}[section]
\newtheorem{corollary}{Corollary}[theorem]
\newtheorem{lemma}[theorem]{Lemma}
\newtheorem{remark}{Remark}

\newcommand{\wei}[1]{\textcolor{purple}{[WP: #1]}}
\newcommand{\avi}[1]{\textcolor{red}{AD: #1}}

\maketitle
\thispagestyle{empty}
\pagestyle{empty}
\setlength{\belowcaptionskip}{-10pt}

\begin{abstract}
Real-world robots must operate under evolving dynamics caused by changing operating conditions, external disturbances, and unmodeled effects. These may appear as gradual drifts, transient fluctuations, or abrupt shifts, demanding real-time adaptation that is robust to short-term variation yet responsive to lasting change. We propose a framework for modeling the nonlinear dynamics of robotic systems that can be updated in real time from streaming data. The method decouples representation learning from online adaptation, using latent representations learned offline to support online closed-form Bayesian updates. To handle evolving conditions, we introduce a changepoint-aware mechanism with a latent variable inferred from data likelihoods that indicates continuity or shift. When continuity is likely, evidence accumulates to refine predictions; when a shift is detected, past information is tempered to enable rapid re-learning. This maintains calibrated uncertainty and supports probabilistic reasoning about transient, gradual, or structural change. We prove that the adaptive regret of the framework grows only \emph{logarithmically in time} and \emph{linearly with the number of shifts}, competitive with an oracle that knows timings of shift. We validate on cartpole simulations and real quadrotor flights with swinging payloads and mid-flight drops, showing improved predictive accuracy, faster recovery, and more accurate closed-loop tracking than relevant baselines.
\end{abstract}


\section{Introduction}

Robotic systems deployed in the real world inevitably face dynamics that deviate from those assumed during design and training~\cite{hersch2008, saviolo2023active,
jia2023evolver, zhou2025simultaneous,  wei2025mlmpcc}. Factors such as payload variation, component wear, environmental disturbances, and unmodeled couplings introduce nonstationary; uncertain, and sometimes abrupt \emph{shifts} --- for example, from payload drops or sudden environmental disturbances. These distribution shifts are particularly challenging in safety-critical domains such as aerial robotics, where failure to adapt rapidly can compromise both performance and safety. A key open question in robotics and learning-based control is therefore how to endow models with the ability to adapt online to evolving dynamics~\cite{zhou2025simultaneous, wei2025mlmpcc, joshi2021asynchronous, mckinnon2021meta, belkhale2021model} while maintaining statistical efficiency, computational tractability, and principled uncertainty quantification.

\begin{figure}[t]
    \centering
    \includegraphics[width=0.48\textwidth]{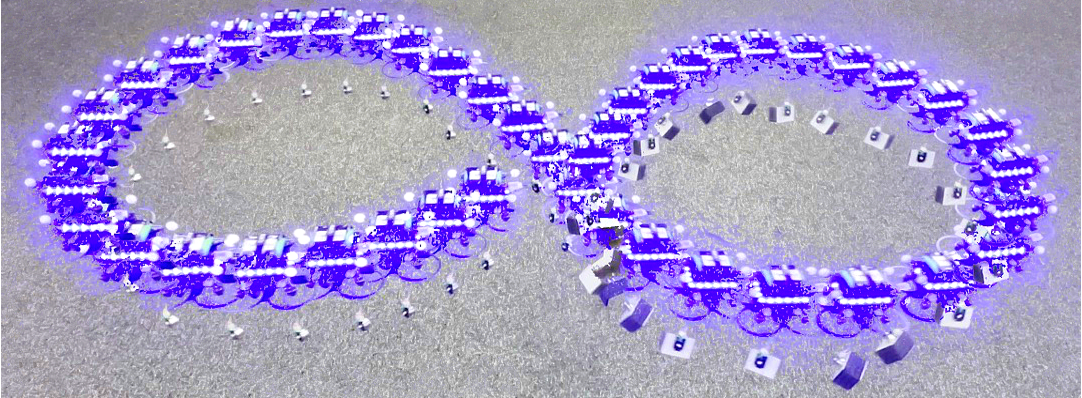}
    \caption{A quadrotor tracks a figure-eight trajectory, starting from the center, to demonstrate its ability to handle payload-induced swing and abrupt release. It carries the payload through the right loop and drops it before entering the left loop.} 
    \label{fig:drone_infinity} 
\end{figure}

\begin{figure*}[t]
\centering
\includegraphics[width=\textwidth]{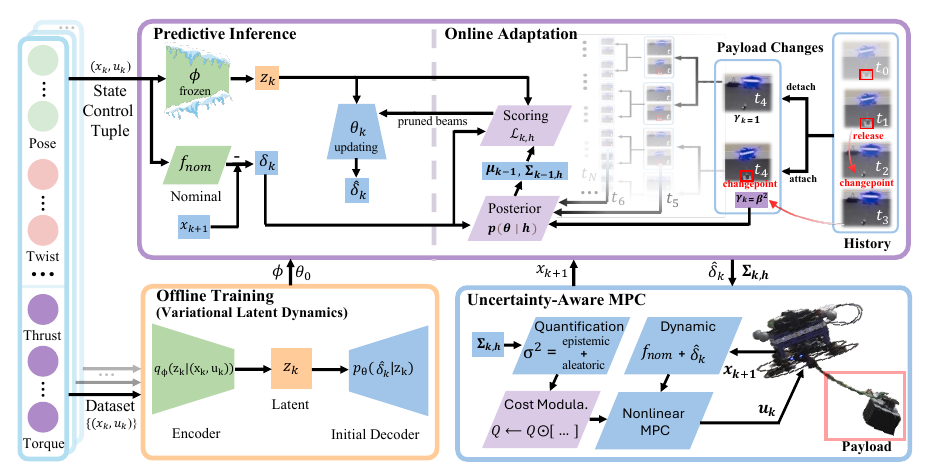}
\label{sec:online}
\caption{\textbf{Overview of online model adaptation and MPC.} \textbf{(A)} We learn unmodeled dynamics \( \delta_k = \theta_k z_k \) on top of nominal dynamics \( f_{\text{nom}}(x_k, u_k) \), where \( z_k \sim q_\phi(z_k \mid x_k, u_k) \) is the encoder parametrized by $\phi$ generating compact system features and the decoder \( \theta_k \) is adapted online. \((\phi, \theta_0)\) is trained on an offline dataset using VLD objective (Section \ref{sec:offline}). \textbf{(B)} At deployment, \( \phi \) is frozen, and \( \theta_k \) is updated using changepoint-aware Bayesian inference, with posterior tempering via \( \gamma_k \in \{1, \beta^2\} \). Each changepoint path \( h \in \mathcal{H}_k \) defines a posterior \( p(\theta \mid h) = \mathcal{N}(\mu_{k,h}, \Sigma_{k,h} / \gamma_k) \), and predictions are obtained by marginalizing over beam-tracked hypotheses weighted by log-likelihood scores \( \mathcal{L}_{k,h} \) (Section \ref{sec:online}). \textbf{(C)} The total predictive variance \( \sigma^2_{\text{tot}}(x_k, u_k) \) captures epistemic (arising from decoder parameters and changepoint path) and aleatoric uncertainty, and is used to modulate the state cost $Q$ of MPC, enabling robust control under regime shifts such  payload release (Section \ref{sec:mpc}).}
\label{pipeline}

\end{figure*}

While several learning-based approaches have explored online adaptation, existing strategies face key limitations in nonstationary settings. Some methods~\cite{jiahao2023online} require full model retraining, which is computationally intensive and ill-suited for streaming or latency-sensitive applications. Others~\cite{wei2025mlmpcc, mckinnon2021meta, o2022neural, gu2024proto, chakrabarty2024physics} rely on interpolation among a fixed set of pre-trained models, restricting generalization to previously unseen regimes. Lightweight alternatives, such as updating the final layer via gradient descent~\cite{saviolo2023active, zhou2025simultaneous,  wei2025mlmpcc}, improve efficiency but exhibit a sharp speed–stability trade-off (large steps adapt fast but can be unstable; small steps are stable but slow after abrupt changes). More importantly, none of these approaches has an explicit mechanism to distinguish between gradual variation and abrupt structural shifts. After a sudden change, past information continues to influence the model, persistently biasing predictions until enough new data accumulate. This leads to a characteristic adaptation lag, during which predictive accuracy and control performance degrade. In safety-critical robotic systems, even short transients of degraded performance can be costly. To mitigate this, a shift-aware mechanism is needed: one that can rapidly reset or re-weight outdated information when evidence of a regime change appears, while still preserving statistical efficiency and calibrated uncertainty during periods of stability.

To address these challenges, we propose a framework that combines shift-aware adaptation with a lightweight and uncertainty-aware model structure. We explicitly decouple representation learning from online adaptation, enabling efficient inference while preserving expressiveness. Offline, we train a nonlinear encoder using a variational objective to learn a structured latent space of dynamics features. Online, we adapt only a lightweight linear decoder on top of this frozen encoder using Bayesian Linear Regression (BLR). This design yields three key advantages: \textit{(i) Data efficiency:} latent features learned offline provide well-conditioned inputs for fast BLR updates, \textit{(ii) Uncertainty quantification:} BLR admits closed-form posterior updates with calibrated variance estimates; and \textit{(iii) Computational tractability:} only a small decoder matrix is adapted at deployment, avoiding encoder retraining.

To handle the reality of distributional shifts, we introduce a changepoint-aware Bayesian update mechanism. At each time step, the model evaluates the likelihood of a regime switch and, when evidence supports a changepoint, it tempers or resets the prior precision, thereby reducing the influence of outdated information. Instead of committing to a single explanation of the data, the method maintains a small set of competing hypotheses over changepoint histories, each with its own decoder posterior. This enables the model to reason probabilistically about whether observed deviations arise from noise, gradual drift, or genuine structural change, and to adapt uncertainty accordingly. As a result, the framework remains responsive to abrupt shifts while allowing rapid re-convergence once new data accumulates.

Beyond algorithmic design, we establish theoretical guarantees that formalize the reliability of our framework. Specifically, we show: (i) posterior consistency in stationary regimes, ensuring that the model converges to the true dynamics; (ii) bounded predictive variance under changepoints, preventing uncertainty blow-up; and (iii) an adaptive regret bound with respect to the best piecewise-stationary model in hindsight, scaling only \emph{logarithmically in time} and \emph{linearly with the number of shifts}. Together, these results demonstrate that our method learns efficiently when conditions are stable, remains well-behaved under abrupt shifts, and \emph{adapts near-optimally} to evolving dynamics.

We integrate our adaptive model into a model predictive control (MPC) framework with uncertainty-aware cost modulation, enabling the controller to adjust its aggressiveness based on predictive confidence. The full pipeline is validated in a two-stage experimental program: (i) controlled cartpole simulations with known disturbances and parametric shifts, and (ii) real quadrotor flights with swinging payloads and abrupt mid-flight drops. Across both settings, our method consistently outperforms state-of-the-art learning-based baselines in terms of predictive accuracy, responsiveness to shifts, and control performance.

In summary, the main contributions of this work are:
\begin{itemize}
    \item A latent-variable framework that decouples offline representation learning from online Bayesian adaptation.
    \item A changepoint-aware Bayesian update mechanism that enables rapid recovery after abrupt shifts while preserving efficiency during stable periods.
    \item Theoretical guarantees covering posterior consistency, bounded predictive variance, and adaptive regret under nonstationarity.
    \item An uncertainty-modulated MPC integration for robust closed-loop control.
\end{itemize}

\subsection{Related Works}
Robust MPC~\cite{mayne2011tube, 10214438} and stochastic MPC~\cite{mesbah2016stochastic, 8909368} account for uncertainty but rely on assumptions, such as boundedness or known distributions, that are often difficult to ascertain in practice. Recent works have advanced learning-based MPC for quadrotors through policy-search integration \cite{9719129}, neural dynamics models for real-time control \cite{10049101}, and learned optimization strategies \cite{10611492}. To improve the predictive accuracy, a range of offline learning methods have been explored, employing Gaussian Processes (GPs) \cite{torrente2021data, cao2024computation}, NeuralODEs \cite{duong2021hamiltonian, chee2022knode, chee2023enhancing}, NeuralSDEs \cite{djeumou2023learn}, Deep Neural Networks (DNNs) \cite{shi2019neural, salzmann2023real}, Physics-Informed Temporal Convolutional Networks \cite{saviolo2022physics} and Diffusion \cite{das2025dronediffusion}. While capable of modeling dynamics beyond analytic formulations, offline models trained on fixed datasets often fail to generalize under deployment-time distribution shifts, leading to compounding prediction errors~\cite{lambert2022investigating} and degraded control performance without online adaptation~\cite{saviolo2023learning}.

Recent work also addresses nonstationary and time-varying disturbances by adapting~\cite{o2022neural, gu2024proto, chakrabarty2024physics} or updating the dynamics model~\cite{saviolo2023active, jia2023evolver, zhou2025simultaneous, wei2025mlmpcc, mckinnon2021meta,  jiahao2023online, gahlawat2020l1, lew2022safe, richards2021adaptive, kaushik2020fast}. Some approaches based on deep neural networks update only the final layer of the model via Gradient Descent~\cite{saviolo2023active, zhou2025simultaneous, wei2025mlmpcc}, which requires careful optimizer tuning and struggles with limited data~\cite{jastrzkebski2017three, smith2017cyclical}. Others periodically retrain the entire network and combine parameters using an exponential moving average \cite{jiahao2023online}, introducing significant computational cost. Meta-Learning methods~\cite{mckinnon2021meta, o2022neural, gu2024proto, chakrabarty2024physics, lew2022safe, richards2021adaptive} interpolate among pre-trained models, restricting generalization to unseen dynamics~\cite{wei2025mlmpcc, kaushik2020fast}. Meanwhile, some approaches~\cite{joshi2021asynchronous, joshi2019deep} adapt only the control policy during deployment, without explicitly updating the underlying dynamics model, which may limit robustness under significant shifts in system behavior or unmodeled dynamics. Although some work update the dynamics model using GP regression~\cite{gahlawat2020l1}, Bayesian last-layer update~\cite{mckinnon2021meta, lew2022safe} and adaptive controller~\cite{richards2021adaptive}, they lack mechanisms to detect and respond to abrupt shifts. In contrast, we consider a more realistic setting, where the dynamics is nonstationary, with potentially abrupt shifts.

Our work adopts a structured system identification approach that separates representation learning from online adaptation, enabling lightweight updates while maintaining predictive uncertainty estimates. This draws on principles from latent-variable modeling for dynamics prediction~\cite{ramos2019bayessim,feng2022factored, wu2023daydreamer}, Bayesian online learning~\cite{nguyen2018variational, kurle2019continual, mania2022active, li2021detecting} and changepoint-aware adaptation strategies~\cite{adams2007bayesian, knoblauch2018doubly, saatcci2010gaussian}. By combining these elements, our framework achieves online streaming adaptation to transient, abrupt, and intermittent shifts in system dynamics, ensuring sustained predictive performance under nonstationary conditions.

\section{Methodology}  
Real-world robotic systems rarely operate under fully known or stationary dynamics, often encountering latent, time-varying, or unmodeled effects. To capture this, we consider the discrete-time system \( x_{k+1} = f(x_k, u_k) \), where \( f \) denotes the true dynamics, and assuming uncertainty is separable from the nominal component~\cite{o2022neural, torrente2021data, shi2019neural}, we decompose the system as:
\begin{equation}
x_{k+1} = f_{\text{nom}}(x_k, u_k) + f_\Delta(x_k, u_k) + \epsilon_k, 
\end{equation}
where $\epsilon_k \sim \mathcal{N}\big(0, \mathrm{diag}(\sigma_1^2, \ldots, \sigma_d^2)\big)$, \( x_k \in \mathbb{R}^d \) is the system state, \( u_k \in \mathbb{R}^m \) is the control input, \( f_{\text{nom}} \) is the nominal dynamics, and \( f_\Delta \) captures lumped model uncertainties. The term \( \epsilon_k \) represents zero-mean Gaussian noise with coordinate-wise variance. We parameterize \( f_\Delta \) to support fast adaptation to nonstationary dynamics at deployment, using a latent-variable model: \( f_\Delta(x_k, u_k) = \theta_k \phi(x_k, u_k) \), where \( \phi: \mathbb{R}^{d+m} \rightarrow \mathbb{R}^\ell \) is a nonlinear encoder trained offline, and \( \theta_k\in\mathbb{R}^{d\times \ell} \) is a linear decoder updated online~\cite{mckinnon2021meta, o2022neural}. This decouples representation learning from adaptation, enabling efficient streaming updates to \( \theta_k \) while avoiding encoder retraining. The linear decoder permits tractable Bayesian updates with uncertainty quantification, while the smooth latent space improves posterior conditioning for data-efficient online learning.

\subsection{Offline Training} \label{sec:offline}
The offline phase aims to learn encoder--decoder pairs that ensure the latent space is expressive and well-conditioned for efficient adaptation of the decoder post-training. Specifically, we learn parameters \( \{ \phi, \theta_0 \} \), where \( \theta_0 \) is trained offline and used to initialize \( \theta_k \) for adaptation at deployment.
To this end, we frame offline training as approximate inference in Variational Latent Dynamics (VLD) model~\cite{wu2023daydreamer, hafner2019learning}. In this setup, we posit a generative process over state transitions conditioned on a latent variable \( z_k \):
\begin{equation}
\label{eqn:vld}
    z_k \sim q_\phi(z_k \mid x_k, u_k), \quad f_{\Delta} \sim \mathcal{N}(\theta_0 z_k, \sigma^2 I),
\end{equation} 
with a standard Gaussian prior on \( z_k \). The training objective is a variational bound on the marginal likelihood of observed transitions:
{\small
\begin{align}
\mathcal{L}_{\text{var}}(\phi, \theta_0) &= \sum_{k=1}^{N} \mathbb{E}_{z_k \sim q_\phi} \left[ \|x_{k+1} - \{ f_{\rm nom}(x_k,u_k) + \theta_0 z_k \} \|^2 \right] \nonumber \\
&+ \beta_{\rm KL} \cdot \mathrm{KL}\left(q_\phi(z_k \mid x_k, u_k) \, \| \, \mathcal{N}(0, I)\right).
\end{align}
}
This approach regularizes the latent space, discourages feature collapse, and improves posterior conditioning. The resulting representations support richer latent geometry and enable stable Bayesian updates during online adaptation. See Appendix~\ref{appendix:posterior_conditioning} for further discussion.

\subsection{Online Adaptation}
\label{sec:online}
While the offline-trained latent model captures system dynamics via learned latent representations, it assumes that deployment-time conditions match the training distribution. In practice, this assumption is often violated due to distributional shifts arising from nonstationary factors such as payload changes, wind, or previously unseen scenarios, leading to model prediction errors and degraded control performance. To address this, we adopt an online adaptation strategy that updates only the linear decoder \( \theta \), while keeping the pre-trained encoder \( \phi \) fixed. This separation ensures that representation learning, which typically requires substantial data and computation, is handled offline, while online updates focus on adapting the decoder to evolving system dynamics in a data-efficient and computationally lightweight manner. Furthermore, the data \( (x_k, u_k, x_{k+1}) \) observed during deployment are not necessarily i.i.d., but rather drawn from a time-varying distribution \( p_k(x_k, u_k, x_{k+1}) \) that evolves in response to unobserved changes in the environment or system configuration.\footnote{The time-varying distribution \( p_k(x_k, u_k, x_{k+1}) \) can be viewed as arising from a latent, possibly non-Markovian process \( \{\eta_k\} \) that governs the system's operating regime. Our framework makes no structural assumptions about \( \{\eta_k\} \), allowing for both abrupt changepoints and gradual drifts. The goal is to infer \( \theta_k \) in a statistically efficient and robust manner under such nonstationarity.} These shifts may occur gradually or abruptly, and we make no assumptions about their timing or smoothness. This setting introduces a central challenge: how to optimally adapt \( \theta_k \) to each new observation while still leveraging relevant information from past data.

\paragraph{Bayesian Formulation}  
We adopt a Bayesian Linear Regression (BLR) formulation for the decoder, as it enables closed-form, uncertainty-aware updates that support efficient online inference, crucial for on-the-fly adaptation. This also provides principled posterior variance estimates, which we later leverage to modulate the control gains during control optimization. Given the fixed encoder \( \phi \), we obtain latent features \( z_k = \mu_\phi(x_k, u_k) \). The decoder is then modeled as a BLR problem, where the dynamics are expressed as:
\begin{equation}
\label{eqn:blr_problem}
x_{k+1} = f_{\rm nom}(x_k, u_k) + \theta_k z_k + \epsilon_k, 
\end{equation}
where $\epsilon_k \sim \mathcal{N}\left(0, \operatorname{diag}(\sigma_1^2, \ldots, \sigma_d^2)\right)$, \( \theta_k \) is the decoder matrix adapted at time step \( k \), and \( z_k \in \mathbb{R}^{\ell} \) is the latent feature vector. Each row \( \theta_k^j \) corresponds to the decoder weights for output dimension \( j \).

A natural question that arises in this latent-to-linear formulation is:\\

\textit{Is online adaptation via the linear decoder alone expressive and flexible enough to capture the true dynamics?}\\

This is not obvious, since the nonlinear encoder $\phi$ is deliberately frozen after offline training while performing online updates only to the linear decoder $\theta_k$. To address this, we consider a simplified scenario in which the system dynamics remains stationary and the true decoder $\theta^*$ is fixed. Under this assumption, the update rule for $\theta_k$ reduces to the classic BLR over latent features. The following lemma shows that, in this setting, the posterior over $\theta_k$ concentrates around the true decoder, and the uncertainty vanishes over time, establishing statistical consistency of the online adaptation process.
\begin{lemma}[Posterior Consistency in Stationary Regimes]
\label{lemma:posterior_consistency}
Suppose the dynamics follow:
\[
x_{k+1}  = f_{\rm nom}(x_k, u_k) + \theta^\star z_k + \epsilon_k, \quad \epsilon_k \sim \mathcal{N}(0, \sigma^2 I),
\]
with $z_k=\phi(x_k, u_k)$ satisfying $\|z_k\| \leq R < \infty$. Let $\theta^{\star j}$ denote the true decoder weights for output dimension $j$, with prior $\theta^j \sim \mathcal{N}(\mu_0, \Sigma_0)$, and let the design matrix be $Z_T = [z_1^\top, \dots, z_T^\top]$. If
\[
\lambda_{\min} \left( \sum_{k=1}^T z_k z_k^\top \right) \to \infty \quad \text{as } T \to \infty,
\]
then for each $j$, the Bayesian posterior mean $\mu^{j}_{(T)}$ converges in mean square to $\theta^{\star j}$, and the posterior covariance $\Sigma^{j}_{(T)} \to 0$:
\[
\mathbb{E}\left[ \|\mu^{j}_{(T)} - \theta^{\star j}\|^2 \right] \to 0, \quad \text{and} \quad \operatorname{tr}(\Sigma^{j}_{(T)}) \to 0.
\]
\end{lemma}
This result provides a crucial theoretical foundation for our framework. It confirms that, under stationary conditions, the online adaptation mechanism is not only efficient but also \emph{statistically consistent}: the decoder learns the correct dynamics without needing to retrain or modify the encoder. In practice, this ensures stable and accurate system identification in environments that are slowly varying or intermittently stable, while setting a baseline for how the method should behave when no distribution shifts occur. The lemma also underlines the importance of offline training: by ensuring the encoder provides sufficiently informative and well-conditioned features, the decoder can leverage them effectively during online learning.

\paragraph{Changepoint-Aware Posterior}  
To remain responsive under nonstationary conditions, the model must distinguish between gradual variation and genuine structural shifts. Without mechanisms to detect or respond to regime shifts, recursive Bayesian updates may propagate outdated information, leading to posterior inconsistency and reduced adaptability. To address this, we introduce a latent binary variable \( c_k \in \{0, 1\} \) at each time step: \( c_k = 0 \) indicates continuity in the system dynamics regime, while \( c_k = 1 \) denotes a changepoint, i.e., a possible abrupt shift. This variable determines how the previous posterior is used as the prior for the current update. When \( c_k = 0 \), standard recursive Bayesian updates are performed, with uncertainty shrinking over time. When \( c_k = 1 \), the prior's influence is downweighted by tempering its precision using a scaling factor \( \gamma_k \),
 defined as:
\begin{align}
\label{eqn:tempering} 
\gamma_k = \begin{cases}
     1, &\text{ if } c_k = 0 \ (\text{no change}); \\
     \beta^2, &\text{ if } c_k = 1 \ (\text{changepoint})
\end{cases}
\end{align}
with \( \beta \in (0, 1) \) acting as a temperature parameter that controls the degree of prior weakening~\cite{fearnhead2007line}.  This strategy ensures that, upon detecting a changepoint, the model remains flexible enough to accommodate novel patterns in the data-generating process, while still leveraging structure from prior experience. It also enables smooth interpolation between full memory retention ($\gamma_k = 1$) and full reset ($\gamma_k \to 0$), providing a principled means to balance robustness and adaptability during online learning. However, introducing such flexibility raises another concern:\\

\textit{Can posterior uncertainty grow without bound if changepoints occur repeatedly or frequently?}\\

This concern is particularly relevant in practical settings where changepoints may occur sporadically or cluster over short time intervals. Without safeguards, the repeated application of posterior tempering could accumulate uncertainty faster than it can be reduced by new observations, resulting in degraded predictive confidence and unstable downstream decision-making. This would undermine the very purpose of Bayesian modeling in this context: providing calibrated uncertainty estimates that inform both learning and control. 

The following lemma formally addresses this issue by showing that the predictive variance remains \emph{uniformly bounded over time}, even in the presence of multiple changepoints. As long as the number of changepoints $\kappa$ is finite and the latent features $z_k$ are bounded, the predictive variance admits a worst-case envelope that depends on the prior scale and segmentwise feature excitation; and when aggregated across segments, grows at most \emph{linearly} with the number of changepoints $\kappa$. 

\begin{lemma}[Bounded Predictive Variance under Changepoint]
\label{lemma:bounded_variance}
    Assume $\|z_k\| \leq R$ for all $k$, and that at most $\kappa$ changepoints occur over $T$ time steps. For output dimension $j$, let the posterior covariance be tempered with $\gamma_k\in\{1, \beta^2\}$ with noise variance $\sigma_j^2>0$. Suppose that, within each stationary segment $s$ of length $T_s$ (i.e., between consecutive changepoints), the minimum eigenvalue of the accumulated feature matrix satisfies 
    \[\lambda_{\min} \left( \sum_{t\in s} z_t z_t^\top \right) \geq \alpha T_s \text{ for some } \alpha>0.\]
    Then, for any time $k$ lying in a segment $s$,
    \[z_k^\top\Sigma_{k,h}^jz_k\leq \min\left\{\tau^2R^2, \frac{R^2\sigma_j^2}{\alpha t(k)}\right\}\leq \tau^2R^2+\frac{R^2\sigma_j^2}{\alpha},\]
    where $t(k)$ denotes the number of samples elapsed in $s$ up to time $k$ (with $t(k)\ge 1$ immediately after a changepoint). In particular, the one-step predictive variance remains uniformly bounded and; the worst-case envelope across at most $\kappa$ resets grows at most linearly with $\kappa$ when aggregated segment-wise.
\end{lemma}
This result guarantees that our model remains statistically well behaved under regime-switching conditions. The predictive variance cannot diverge: it is controlled by the prior scale $\tau^2$, the feature bound $R$, the noise level $\sigma_j^2$, and the segment-wise excitation constant $\alpha$, with at most \emph{linear} dependence on the total number of changepoints $\kappa$. Intuitively, although a changepoint resets the effective prior precision (via $\gamma_k$), subsequent data within the new stationary segment re-accumulates information and contracts the posterior at a $1/t$ rate. Consequently, the model \emph{does not overcommit} to stale posteriors after abrupt changes and avoids variance growth beyond a bounded polynomial envelope across repeated resets.

Each sequence of changepoint decisions $c_{1:k} = (c_1, \dots, c_k)$ induces a distinct trajectory of decoder posteriors. To manage the combinatorial space of possible changepoint paths, we maintain a beam $\mathcal{H}_k$ of the top-$K$ hypotheses~\cite{harvey1995limited}, each representing a \emph{unique changepoint history}, and ranked by their joint log-likelihood. 

Assuming a Gaussian prior over the decoder with offline-trained weights and variance $\tau^2$ (i.e., $\theta^j \sim \mathcal{N}(\theta_0^j, \tau^2 I)$), the prior at time $k$ for each hypothesis $h \in \mathcal{H}_k$ is defined as: 
\begin{equation*}
    p(\theta^j \mid h) = \mathcal{N}\fbr{\mu^{j}_{k-1,h}, \frac{1}{\gamma_k}\Sigma^{j}_{k-1,h} }, \quad \forall j = 1, \dots, d
\end{equation*}
The corresponding posterior is recursively updated using the residual \( \delta_{k+1} = x_{k+1} - f_{\text{nom}}(x_k, u_k) \) and latent feature $z_k$ as:
\begin{align}
\label{eqn:posterior_covariance}
\begin{split}
    \Sigma^{j}_{k,h} &= \left( \gamma_k (\Sigma^{j}_{k-1,h})^{-1} + \frac{1}{\sigma_j^2} z_k z_k^\top \right)^{-1} \hspace{-4mm}; \\
\mu^{j}_{k,h} &= \Sigma^{j}_{k,h} \left( \gamma_k (\Sigma^{j}_{k-1,h})^{-1} \mu^{j}_{k-1,h} + \frac{1}{\sigma_j^2} z_k \delta_{k+1}^j \right).
\end{split}
\end{align}
\paragraph{Marginal Likelihood and Hypothesis Scoring}  
The following lemma formalizes the scoring mechanism used in our beam-tracked, changepoint-aware adaptation strategy (see Appendix \ref{app:changepoint_scoring} for derivations). At each time step $k$, the model considers two possible transitions for each beam hypothesis: $c_k = 0$ (no changepoint) and $c_k = 1$ (changepoint). This produces $2K$ candidate hypotheses in total, from which the top-$K$ are retained based on their cumulative log-likelihood scores. Each hypothesis $h\in\mathcal{H}_k$ corresponds to a unique trajectory of changepoints $c_{1:k}$, and maintains its own posterior over decoder weights. The joint score for a hypothesis captures both how well it \emph{explains the observed residuals} and \emph{how plausible its changepoint decisions are}.

\begin{lemma}
\label{lemma:changepoint_scoring}
Let $c_k \in \{0, 1\}$ be the latent changepoint variable at time $k$, $\pi \in(0,1)$ denote the prior changepoint probability. Then, the cumulative log-likelihood of hypothesis $h\in\mathcal{H}_k$ is:
\begin{multline}\label{eqn:likelihood_hypo_recursive}
    \mathcal{L}_{k,h} = \mathcal{L}_{k-1,h} + \log p(\delta_k \mid z_k, c_k) \\
    + \log p(c_k \mid c_{1:k-1}, x_{1:k}, u_{1:k}),
\end{multline}
where the marginal log-likelihood of the residual $\delta_k$ is given by:
{\small
\begin{align}
\log p(\delta_k \mid z_k, c_k) 
&= -\frac{1}{2} \sum_{j=1}^d \Bigg[
    \log\bigg( 2\pi \left( \frac{1}{\gamma_k} z_k^\top \Sigma^{j}_{k-1,h} z_k + \sigma_j^2 \right) \bigg) \notag \\
&\qquad + \frac{ \left( \delta_k^j - z_k^\top \mu^{j}_{k-1,h} \right)^2 }
             { \frac{1}{\gamma_k} z_k^\top \Sigma^{j}_{k-1,h} z_k + \sigma_j^2 }
\Bigg], \label{eqn:marginal_residual}
\end{align}
}
and the posterior probability of a changepoint is computed via Bayes' rule:
{\small
\begin{align}
    p(c_k = 1 \mid \cdot) 
&= \frac{ \pi \cdot p(\delta_k \mid z_k, c_k = 1) }
        { 
        \begin{aligned}
        &\big[\pi \cdot p(\delta_k \mid z_k, c_k = 1) \\
        &\quad + (1 - \pi) \cdot p(\delta_k \mid z_k, c_k = 0)\big]
        \end{aligned}
        } \label{eqn:changepoint_posterior}
\end{align}
}
\end{lemma}
This scoring formulation balances two key components:
\begin{itemize}
    \item \textit{Data fit:} how well the hypothesis explains the observed residual at time $k$, given its belief about whether a changepoint occurred;
    \item \textit{Model prior:} the plausibility of the changepoint event itself under the assumed prior $\pi$.
\end{itemize}

Importantly, changepoint posterior $p(c_k = 1 \mid \cdot)$ plays a central role in adapting the confidence of the model in the current regime. It quantifies the relative evidence for regime change, comparing how well the residual is explained under each scenario ($c_k=0$ vs. $c_k=1$). This can be interpreted as a \emph{soft likelihood ratio test}, where the decisions are not binary but probabilistic, with smoother transitions and graceful fallback to prior knowledge (see Appendix \ref{app:likelihood-ratio_discussion} for further discussion). 

Unlike hard changepoint detection rules (e.g., based on residual magnitude thresholds), this probabilistic framework naturally balances adaptation and stability. It allows the model to remain cautious in the face of transient noise, while still responding decisively to genuine structural changes when supported by sufficient evidence. This is particularly important in robotics and control, where abrupt resets based on unreliable evidence can degrade performance.

\paragraph{Predictive Inference}  
During deployment, the model generates predictions by marginalizing over the current beam of changepoint hypotheses. Each hypothesis $h\in \mathcal{H}_k$ corresponds to a distinct changepoint trajectory and maintains its own posterior over decoder parameters. Given a new latent input $z_*$, the predictive distribution under hypothesis $h$ is a multivariate Gaussian with conditionally independent output dimensions:
\begin{equation*}
    p(\delta_* \mid z_*, h) = \prod_{j=1}^d \mathcal{N}\left(\delta_*^j \mid z_*^\top \mu^{j}_{k,h},\ z_*^\top \Sigma^{j}_{k,h} z_* + \sigma_j^2\right)
\end{equation*}
To account for uncertainty over changepoint histories, we marginalize over the beam, using log-likelihood based weights:
\begin{align}
\label{eqn:prediction}
    p(\delta_* \mid z_*, \mathcal{D}_{1:k}) &= \sum_{h \in \mathcal{H}_k} w_h \cdot p(\delta_* \mid z_*, h), \nonumber \\
    w_h &= \frac{\exp(\mathcal{L}_{k,h})}{\sum_{h' \in \mathcal{H}_k} \exp(\mathcal{L}_{k,h'})}.
\end{align}
This inference mechanism captures two distinct forms of uncertainty:
\begin{itemize}
    \item \textit{Epistemic uncertainty} over the decoder weights, captured by the posterior covariance $\Sigma^j_{k,h}$ within each hypothesis via Bayesian Linear Regression.
    \item \textit{Structural uncertainty} over the true changepoint path, captured by the beam-based marginalization over $\mathcal{H}_k$.
\end{itemize}

\paragraph{Adaptive Regret under Nonstationarity}
In dynamic environments where system behavior evolves over time, a key challenge is to maintain accurate predictions without overreacting to noise or transient variation. Ideally, a model should adapt quickly to regime shifts while retaining relevant information when the dynamics remain stable. This raises the following question:\\

\textit{How well do we compete against a benchmark that has access to the underlying regime structure in hindsight, without knowing the changepoint structure in advance?}\\

To formalize this, we consider a standard comparator class: the set of linear decoders that are piecewise constant with a bounded number of changepoints. This class serves as a structured benchmark for analysis and is not assumed or required by the model itself. The following theorem provides a regret bound with respect to this class.

\begin{theorem}[Adaptive Regret Bound under Piecewise Stationarity]
\label{thm:regret}
Let the true data-generating process be piecewise stationary, with at most \( \kappa \) changepoints over a horizon \( T \). Let \( \hat{\delta}_k \) denote the predictive mean under the model, and let \( \mathcal{G}_\kappa \) be the class of linear decoders that are piecewise constant with at most \( \kappa \) segments. Then the cumulative squared prediction error satisfies:

\begin{align*}
    \sum_{k=1}^T \mathbb{E}\left[ \| \delta_k - \hat{\delta}_k \|^2 \right] &- \min_{\theta \in \mathcal{G}_\kappa} \sum_{k=1}^T \| \delta_k - \theta_{[k]} z_k \|^2 \\
    &= \mathcal{O}(\log T + \kappa \ell \log T),
\end{align*}
\end{theorem}
This bound is \emph{non-asymptotic} and holds uniformly over all sequences with at most $\kappa$ changepoints. It quantifies the worst-case overhead incurred by our method relative to the best piecewise-stationary sequence of decoders in hindsight, in a setting where both the \emph{changepoint locations and regime-specific decoder parameters are latent} and must be \emph{inferred online}. Specifically, the regret scales logarithmically with the time horizon $T$ and linearly with both the number of changepoints $\kappa$ and the latent dimension $\ell$. The logarithmic dependence on $T$ implies that the model \emph{does not suffer cumulative error in stable regimes}; its performance improves over time as more data is observed. Meanwhile, the linear dependence on $\kappa$ reflects that the \emph{cost of adaptation grows only proportionally with the number of regime shifts}, without compounding across time. Together, these scaling properties highlight the method's ability to retain information when the system is stationary, while remaining responsive to nonstationary changes.

\begin{remark}
    The piecewise stationarity assumption is used solely for tractability of analysis, not as a requirement for the algorithm itself. Our method does not assume or require piecewise stationarity during training or deployment. It is designed to handle fully nonstationary dynamics including gradual drifts and noise-driven variations, without relying on any structural assumptions about changepoint frequency, spacing, or smoothness. The theorem shows that even in the presence of latent, unknown, and possibly abrupt regime changes, our method achieves near-optimal performance, without access to oracle information about when or how the dynamics change.
\end{remark}

\section{Uncertainty-Aware Cost Modulation in MPC}
\label{sec:mpc}
\begin{algorithm}[h]
\caption{Online Bayesian Adaptation with Changepoint-Aware MPC}
\label{alg:online_adaptation}
\KwIn{Pre-trained VLD weights $(\phi, \theta_0)$, prior variance $\tau^2$, noise variances $\{\sigma_j^2\}$, changepoint prior $\pi$, temperature $\beta$, beam size $K$}
\KwOut{Final posterior ensemble $\mathcal{H}_T$}

Initialize posterior mean $\mu_0^j \gets \theta_0^j$, covariance $\Sigma_0^j \gets \tau^2 I$ for all $j = 1,\dots,d$\;
Initialize beam $\mathcal{H}_0 \gets \{ (\mu_0, \Sigma_0, \mathcal{L}_0 = 0, c_{1:0} = \emptyset) \}$\;

\For{$k = 1$ \KwTo $T$}{
    \tcc{Predict}
    Observe $(x_k, u_k)$ and compute $z_k = \mu_\phi(x_k, u_k)$\;
    Compute predictive mean $\hat{\delta}_k$ and variance via \eqref{eqn:prediction}\;
    
    \tcc{Control}
    Generate control input using MPC

    \tcc{Observe new data}
    Observe $x_{k+1}$ and compute $\delta_k = x_{k+1} - f_{\rm nom}(x_k, u_k)$\;
    
    Initialize candidate set $\mathcal{C}_k \gets \emptyset$\;
    \ForEach{hypothesis $h \in \mathcal{H}_{k-1}$}{
        Compute $\log p(\delta_k \mid z_k, c_k = 0)$ and $\log p(\delta_k \mid z_k, c_k = 1)$ via \eqref{eqn:marginal_residual}\;
        Compute changepoint posterior $p(c_k = 1 \mid \cdot)$ via \eqref{eqn:changepoint_posterior}\;

        \For{$c_k \in \{0, 1\}$}{
            Set $\gamma_k = 1$ if $c_k = 0$, else $\beta^2$ \;
            Update posterior $(\mu_{k,h}^j, \Sigma_{k,h}^j)$ for all $j$ using \eqref{eqn:posterior_covariance}\;
            $\log p(c_k) \gets \log\big(c_k \cdot p(c_k = 1 \mid \cdot)$ \\
            \Indp $ + (1 - c_k) \cdot (1 - p(c_k = 1 \mid \cdot))$ \big)\;
            \Indm
            Update hypothesis score $\mathcal{L}_{k,h}$ using \eqref{eqn:likelihood_hypo_recursive} and add to $\mathcal{C}_k$\;
        }
    }
    Prune beam: keep top-$K$ hypotheses in $\mathcal{H}_k$ by $\mathcal{L}_{k,h}$\;
}
\end{algorithm}

Finally, to achieve robust closed-loop trajectory tracking, we integrate our adaptive dynamics model into a model predictive control (MPC) framework that explicitly accounts for predictive uncertainty. During execution, control inputs are computed in real time using a nonlinear MPC controller~\cite{gu2024proto}. Specifically, at each time step $k$, the controller solves a finite-horizon optimal control problem (OCP) of the form:
\begin{align}
\min_{\{x_{k+i}, u_{k+i}\}} \quad  J(x_{k:k+N}, u_{k:k+N-1}) &:= \sum_{i=0}^{N-1} J_s(x_{k+i}, u_{k+i}) \nonumber \\
&+ J_f(x_{k+N})  \\
\text{s.t.} \quad  x_{k+i+1} = f(x_{k+i}, u_{k+i}), &\quad x_{k+i} \in \mathcal{X},\ u_{k+i} \in \mathcal{U}, \nonumber \\
 x_0 = x_k, \quad x_N \in \mathcal{X}_f&, \quad \forall i = 0,\dots,N-1 \nonumber
\end{align}
where \( J_s(x_{k+i}, u_{k+i}) = \|x_{k+i} - x_{k+i}^r\|_Q^2 + \|u_{k+i} - u_{k+i}^r\|_R^2 \) is the stage cost, and \( J_f(x_{k+N}) = \|x_{k+N} - x_{k+N}^r\|_P^2 \) is the terminal cost. The reference trajectories \( x_{k+i}^r \), \( u_{k+i}^r \) are given, and \( Q \), \( R \), and \( P \) are positive semi-definite weighting matrices. At each control step $k$, the first control input \( u_{k}^\star \) is executed, and the OCP is re-solved in a receding horizon manner using the updated state estimate.

To enhance robustness under model uncertainty, the state cost matrix \( Q \) is modulated as follows:
{\small
\begin{align}
    Q_{k}
\;&\leftarrow\;
Q_{k}\;\odot\;
\Bigl[
     1 + \alpha_{1}\,
         \log\!\bigl(1+\alpha_{2}\,
                      \sigma_{\text{tot}}^{2}(x_{k},u_{k})\bigr)
\Bigr], ~~\alpha_{1},\alpha_{2}>0. \nonumber \\
\sigma_{\text{tot},j}^{2}&(x_k,u_k)
    = 
    (\sigma_{z}^{2})^{\!\top}
         \Bigl(\textstyle\sum_{h}w_h\,
               \mu_{k,h}^{j}\odot\mu_{k,h}^{j}\Bigr) \nonumber \\      
        &+ \Bigl[\underbrace{
           \textstyle\sum_{h}w_h\,
           \bigl(z_k^{\!\top}\Sigma_{k,h}^{j}z_k\bigr)}
           _{\text{within-beam (parameter) variance}}
          +
         \textstyle\sum_{h}w_h\,
           (\,z_k^{\!\top}\mu_{k,h}^{j}-\bar m_{j})^{2}
       \Bigr] 
\end{align}}
\hspace{-0.2cm}where $\bar m_{j}\!=\!\sum_{h}w_h\,z^{\!\top}\mu_{k,h}^{j},$ and $\sigma^2_z$ is the encoder variance and $\sigma^2_{\text{tot}}(x_k,u_k)$ denotes the total predictive variance, i.e. epistemic uncertainty of the model and aleatoric perception noise from the encoder. This encourages the controller to avoid regions where the dynamics model is uncertain or the encoder is unreliable. Algorithm \ref{alg:online_adaptation} summarizes the closed-loop control, and Appendix \ref{app:hyperparam_selection} provides practical considerations for selecting the associated hyperparameters.

\section{Experiments and Results}
We evaluate our framework in a two-stage experimental program that progresses from a controlled simulated system to a complex aerial robot. The first stage uses a cartpole simulation as a diagnostic testbed, allowing precise injection of disturbances and parameter changes to validate \textit{(i)} the predictive quality of the offline-trained model under stationary conditions, and \textit{(ii)} the predictive quality of the online updated model and the responsiveness of the adaptation mechanism under controlled regime shifts. The second stage transitions to a real quadrotor platform, where the same framework is tested under realistic sensing, actuation, and environmental uncertainties. Here, we assess both model accuracy in open-loop prediction and control performance in a challenging payload disturbance scenario.

Across all experiments, we compare against established baselines (see Appendix \ref{app:baseline} for details), including $\mathcal{L}_1$-adaptive control~\cite{wu2023l1}, supervised MLPs~\cite{saviolo2023active}, Gaussian processes (GP)~\cite{torrente2021data}, Proto-MPC~\cite{gu2024proto}, and classical system identification (SysID)~\cite{brunton2016discovering, kaiser2018sparse}, trained and evaluated under identical conditions across 10 independent trials.

\begin{figure}[h]
  \centering
  \subfloat[Prediction error under stationary dynamics.\label{fig:offline_result}]{
    \includegraphics[width=0.94\linewidth]{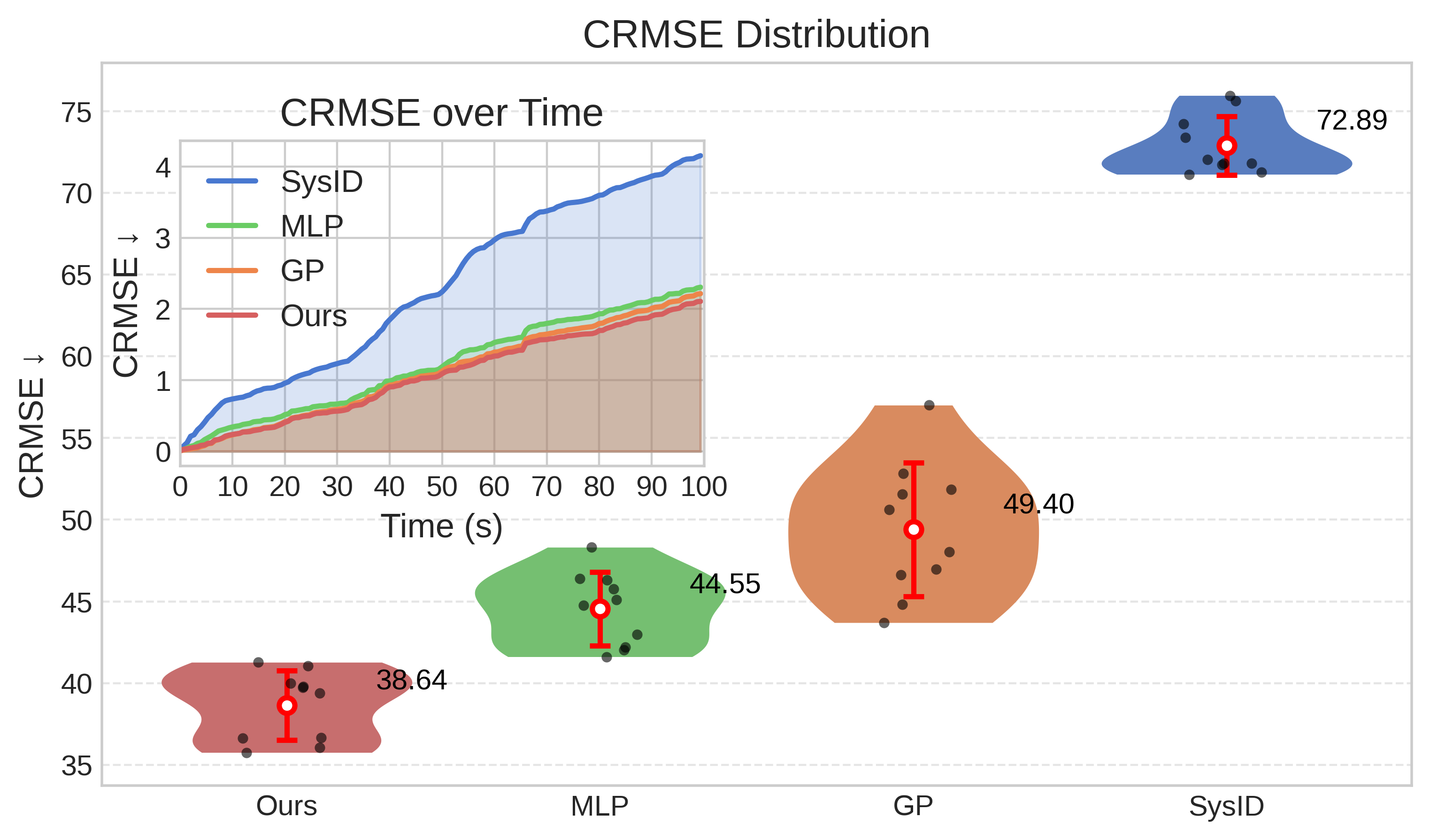}
  }
  \par\medskip

  \subfloat[Prediction error under nonstationary dynamics.\label{fig:online_result}]{
    \includegraphics[width=0.94\linewidth]{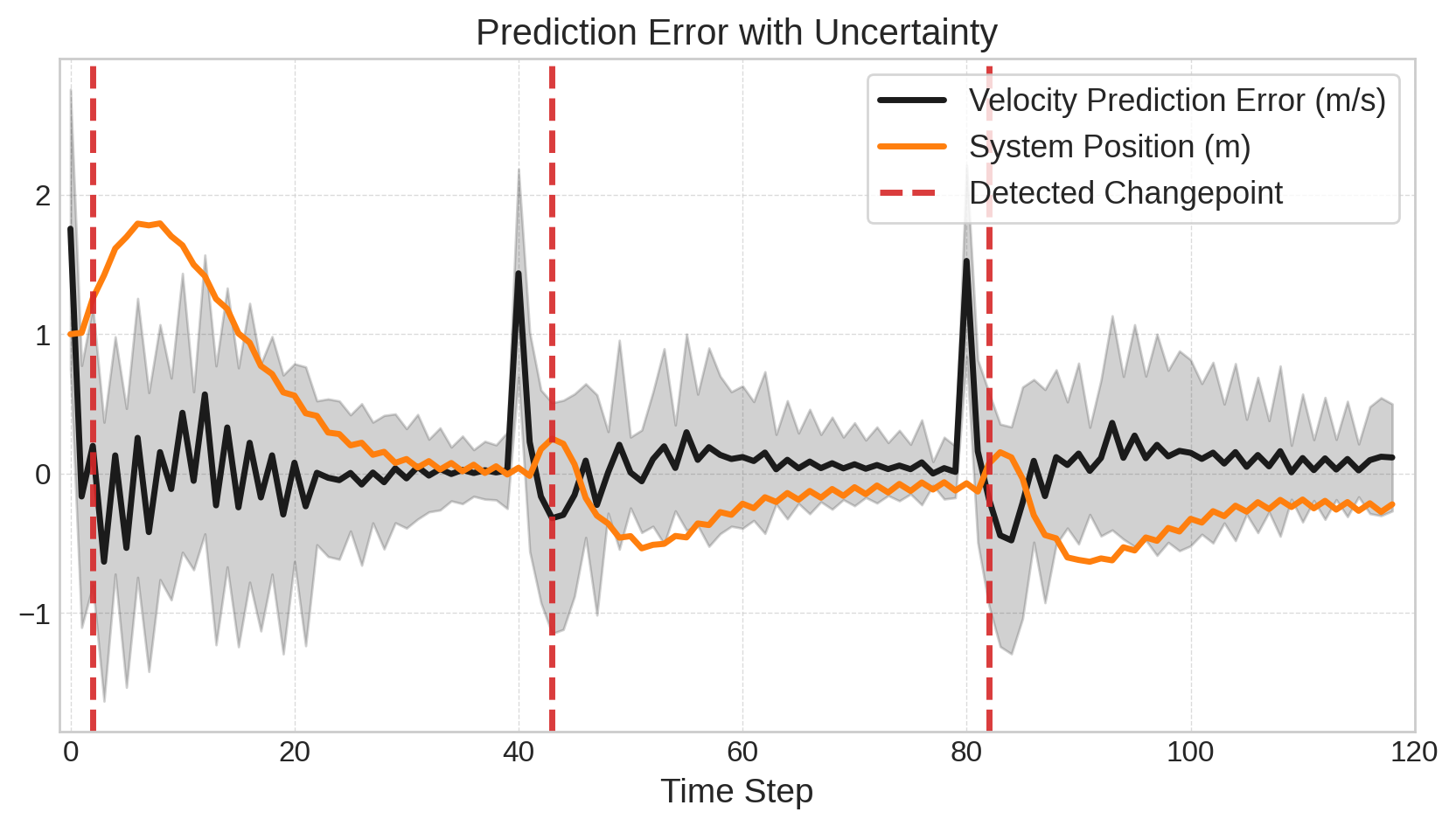}
  }
  \caption{\textbf{Cartpole evaluation.} 
\textit{Top:} Offline predictive accuracy under stationary dynamics. 
Main panel: distribution of CRMSE over the entire test dataset (20 trajectories) for each of the 10 random train/test splits, showing our method achieves the lowest error with a non-overlapping distribution compared to all baselines. 
Inset: time evolution of CRMSE on a single representative test trajectory, where our method shows broadly comparable but lower error growth to most baselines, with a clearer distinction over SysID. 
\textit{Bottom:} Representative online rollout with induced disturbances and mass changes. 
Red dashed lines: detected changepoints; shaded regions: predictive uncertainty from the Bayesian linear update.}
\label{fig:cartpole_results}
\end{figure}
\vspace{-2mm}
\subsection{Cartpole: Controlled Validation and Adaptation Tests}

\subsubsection{Offline Model Validation}\label{sec:offline_validation}
As the first step, we assess the predictive accuracy of the model in a stationary setting without online updates. This tests whether the learned latent-variable model provides both a reliable predictor under fixed dynamics and a strong initialization for subsequent online adaptation. We use a dataset of 100 trajectories from a simulated cartpole system (see Appendix~\ref{app:cartpole}), with 80 used for training and 20 for testing per trial, repeated over 10 random train/test partitions. Performance is measured using Cumulative Root Mean Square Error (CRMSE)\footnote{$\textrm{CRMSE}(x_T,\hat{x}_T)=\sum_{k=1}^T\sqrt{\frac{1}{D}\sum_{j=1}^D(x_k^j-\hat{x}_k^j)^2}$; $D$ = state dimension; $T$ = number of control steps.}. Figure~\ref{fig:offline_result} shows two complementary views: \textit{(i)} the main panel is the distribution of CRMSE over the entire test dataset (20 trajectories) for each of the 10 random splits, where our method achieves the lowest mean error with a non-overlapping distribution compared to all baselines, and \textit{(ii)} the inset shows the time evolution of CRMSE on a single representative test trajectory, where our method exhibits broadly comparable but lower error growth to most baselines, with a clearer distinction over SysID. These results indicate that the offline-trained model is well calibrated under stationary conditions and provides a solid baseline for subsequent online adaptation.

\subsubsection{Online Adaptation under Nonstationary Dynamics}
To assess the effectiveness of the changepoint-aware adaptation mechanism, we evaluate the models in a controlled nonstationary cartpole setting where the timing and nature of the shifts are known. Ten disturbed trajectories of 120 time steps are collected, with interventions at steps 0, 40, and 80: a 2\,N lateral impulse applied to the cart, and a 10\% reduction in both cart and pole masses. These changes induce both exogenous disturbances and parametric shifts. All models are initialized from the offline training described in Section~\ref{sec:offline_validation} and updated online during rollout.  

Table~\ref{tab:online_adaptation} reports CRMSE across trials. Our method achieves the lowest error, with marked degradation when changepoint detection is disabled (Ours w/o CP), underscoring the importance of posterior resets at detected shifts. GP and MLP baselines perform worse, while the offline-only model performs worst, highlighting the benefit of online adaptation. Figure~\ref{fig:online_result} shows a representative rollout: detected changepoints align closely with intervention times, triggering rapid uncertainty resets. Predictive uncertainty rises immediately after each shift and decays as the model adapts, illustrating effective epistemic uncertainty management during online inference.

\begin{table}[h]
\footnotesize
\renewcommand{\arraystretch}{1.1}
\captionsetup{font=footnotesize}
\caption{Online Prediction error on Cartpole}
\centering
\scalebox{0.68}{
\begin{tabular}{lccccc}
\toprule
\textbf{Method} & \textbf{Ours} & \textbf{Ours (w/o CP)} & \textbf{GP}  & \textbf{MLP (last layer)} & \textbf{Ours (offline only)} \\
\midrule
\textbf{CRMSE} & \textbf{5.83 $\pm$ 1.18} & 10.28 $\pm$ 2.07 & 12.53 $\pm$ 3.01 & 17.76 $\pm$ 1.99 &  24.03 $\pm$ 2.34 \\
\bottomrule
\end{tabular}
}
\label{tab:online_adaptation}
\end{table}

\subsection{Real-World Quadcopter Evaluation}
Following the controlled validation in simulation, we evaluate the proposed framework on a real quadrotor platform (see Appendix~\ref{app:quad_details}). This stage serves two purposes: \textit{(i)} open-loop validation of the model with online updates on real flight data, together with a sanity check of key modeling assumptions; and  \textit{(ii)} closed-loop control evaluation in a disturbance-rich scenario with both continuous and abrupt dynamics changes.

\subsubsection{Open-Loop Model Validation and Sanity Check} We train all methods on expert-flown trajectories with fixed payloads of $\{10, 100, 200\}$\,g (3\,min each; 9\,min total). Validation is performed on \emph{disjoint} expert-flown trajectories carrying payloads $\{75,125,175\}$\,g (1\,min each). During validation, models are updated \emph{online} according to their respective update rules (no control is applied). We report CRMSE over each validation trajectory in Table~\ref{tab:quad_model_pred}. To assess our modeling assumptions on real flight data, we monitor Gramian-based metrics derived from $z_i$. The Gramian is defined as $G_t=\sum_{i \le t} z_i z_i^\top$; and we track over a test trajectory: the minimum eigenvalue $\lambda_{\min}(G_t)$ (persistent excitation), the condition number $\kappa(G_t) = \lambda_{\max}(G_t) / \lambda_{\min}(G_t)$ (feature-space conditioning), and the log-determinant $\log \det(G_t)$ (cumulative information volume). We further compute a sliding-window Gramian $G_t^{(W)} = \sum_{i = t-W+1}^{t} z_i z_i^\top$ and track $\lambda_{\min}(G_t^{(W)})$ for $W \in \{15,30\}$ to quantify short-horizon excitation.

\begin{table}[h]
\renewcommand{\arraystretch}{1.1}
\captionsetup{font=footnotesize}
\caption{Open-Loop Prediction Error on Real Quadrotor Flights}
\centering
\scalebox{0.9}{
\begin{tabular}{lccccc}
\toprule
\textbf{Method} & \textbf{Ours} & \textbf{Proto-MPC} & \textbf{MLP} & \textbf{GP}  & \textbf{SysID}  \\
\midrule
\textbf{CRMSE} & \textbf{5103} & 5152 & 6367 & 6804 & 9863 \\
\bottomrule
\end{tabular}
}
\label{tab:quad_model_pred}
\end{table}
\begin{figure}[h]
  \centering
  \subfloat[][Log minimum eigenvalue and sliding-window variants.\protect\\
              \label{fig:eig_value}]{
    \includegraphics[width=0.47\columnwidth]{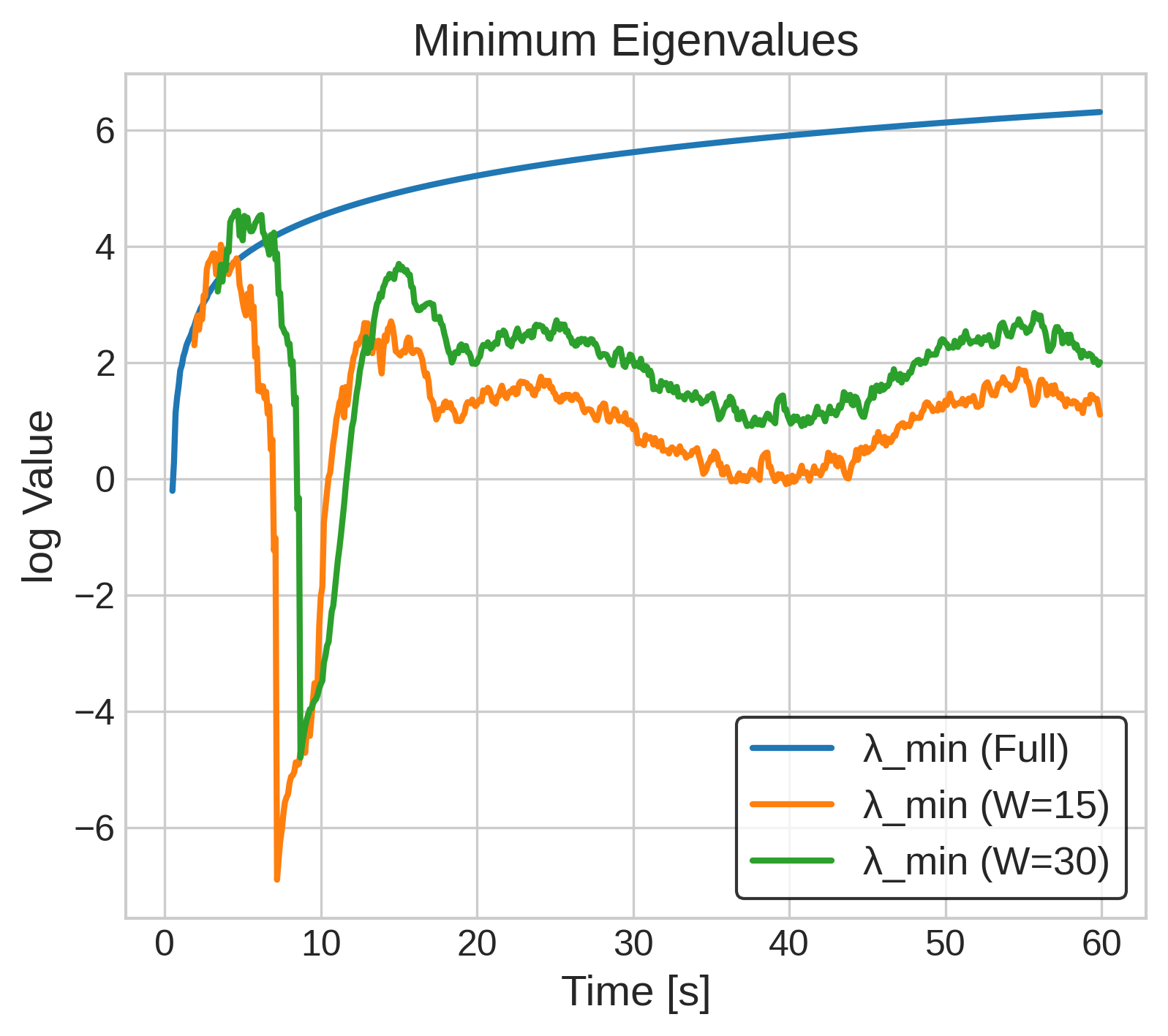}
  }
  \hfill
  \subfloat[][Condition number and log-determinant.\protect\\
              \label{fig:cond_numb}]{
    \includegraphics[width=0.47\columnwidth]{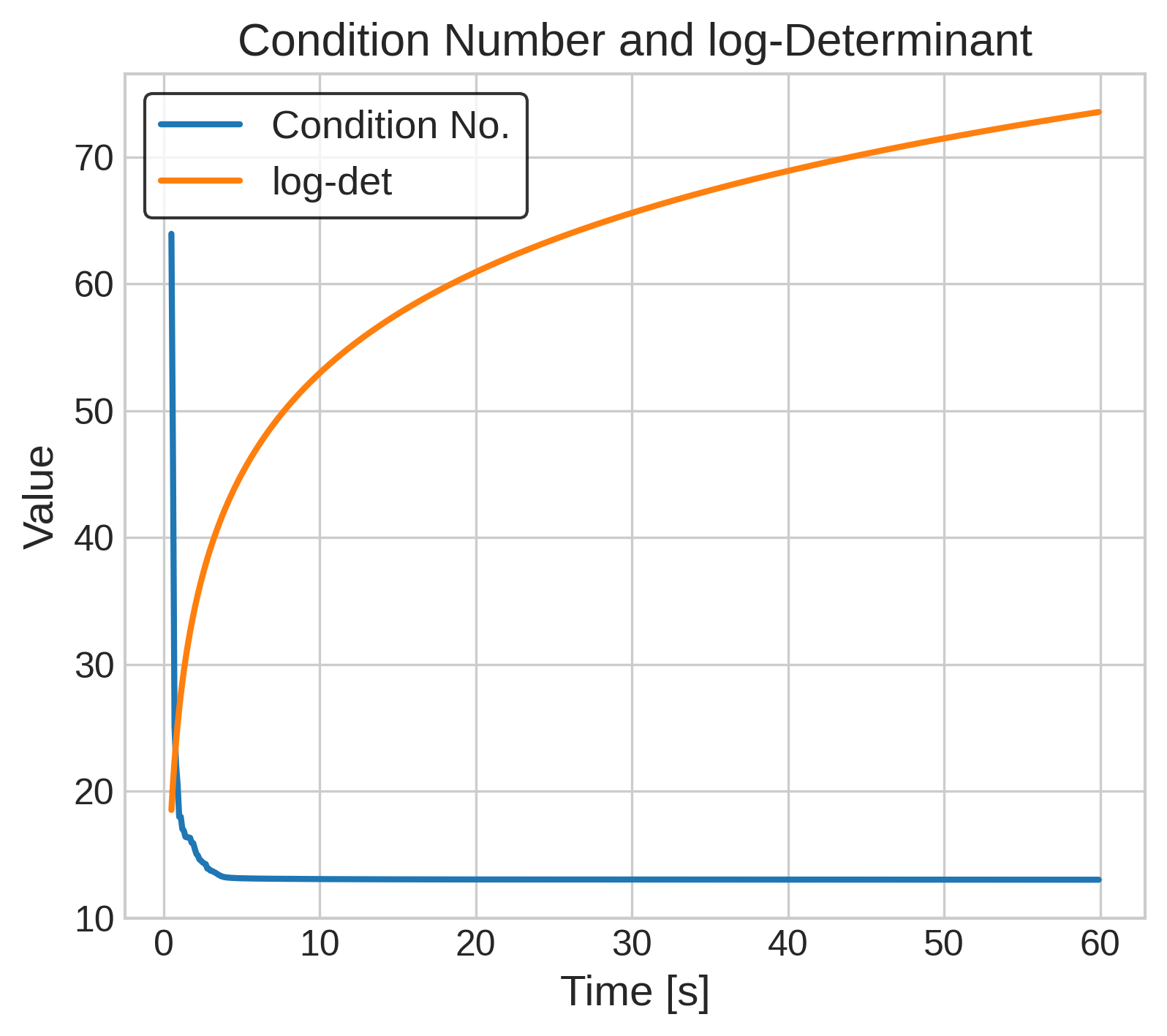}
  }
  \caption{\textbf{Sanity-check metrics on a $1$-minute quadrotor validation trajectory with a $125$\,g payload.} 
  Both plots are computed from encoder outputs $z_t$ and confirm persistent excitation, stable conditioning, and expanding information volume confirming favorable conditions for stable online adaptation.}
  \label{fig:pe_cond}
\end{figure}
Figure~\ref{fig:pe_cond} summarizes these metrics over a $1$-minute validation trajectory. The log minimum eigenvalue $\log\lambda_{\min}(G_t)$ increases steadily and reaches $\approx 6.1$, indicating persistent excitation and a Gramian that grows proportionally with $t$ in all directions (desirable finite-time behavior). The sliding-window curves $\lambda_{\min}(W=15,30)$ exhibit minor dips followed by recovery, confirming local excitation rather than isolated bursts. The condition number $\kappa(G_t)$ starts high, consistent with a train/test mismatch due to different payloads, but rapidly saturates around $12$–$13$; implying that the least-excited latent direction retains roughly $1/13$ the variance of the most-excited one. Thus, the latent geometry remains reasonably isotropic and well-balanced. Finally, $\log\det(G_t)$ grows monotonically, showing that the latent subspace volume expands over time and does not collapse toward a low-rank manifold. Collectively, these trends suggest that online updates are numerically well-posed (bounded step sizes, non-degenerate posterior covariance) and that the learned representation generalizes across payload-induced shifts without incurring ill-conditioning. 

Together, these trends indicate that the learned representation remains well-conditioned and persistently excited in real flight, ensuring that online updates are numerically well-posed. Consistent with this, our method also achieves the lowest predictive error across all tested payloads in Table~\ref{tab:quad_model_pred}, confirming that the framework not only satisfies the theoretical conditions for stable adaptation but also delivers superior predictive performance on real quadrotor dynamics.

\subsubsection{Closed-Loop Nonstationary Scenario}
We now evaluate the complete framework in closed-loop control using MPC. The quadrotor follows a figure-eight trajectory while carrying a swinging payload suspended by a lightweight cable. The oscillatory motion of the payload induces state-dependent, dynamically coupled, and time-varying disturbances with delayed effects. Just before the second lap, the payload is released mid-flight, producing an abrupt change in the thrust-to-mass ratio and an associated transient dynamics shift. This combination of continuous underactuated disturbances (from the swing) and discrete regime changes (from the drop) creates a challenging benchmark for rapid changepoint detection and adaptation in real flight.
\begin{figure*}[t]
\centering
\includegraphics[width=\textwidth]{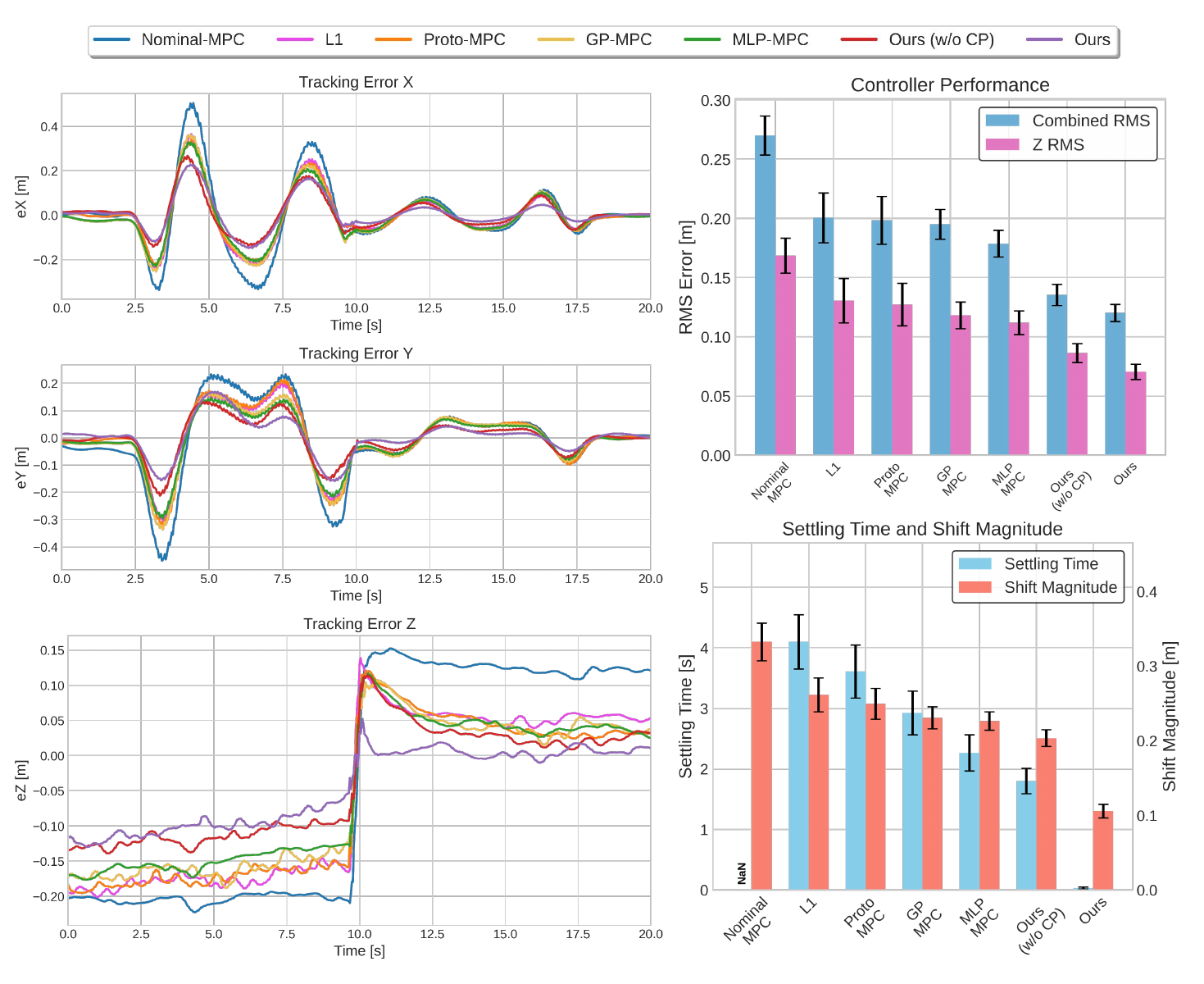}
\caption{
Quadrotor tracking performance comparison for a $175$\,g payload at $2.5$\,m/s. \textbf{\textit{Left:}} Top to bottom plots show trajectory tracking errors in the $x$, $y$, and $z$ coordinates respectively. \textbf{\textit{Right:}} The bar plots report metrics averaged over $10$ trials. The top plot shows combined RMS and $z$-axis RMS errors (accounting for the majority of the combined error). The bottom plot compares shift magnitude and settling time, quantifying abrupt disturbance sensitivity and recovery performance. \emph{shift magnitude} is defined as the absolute change in tracking error immediately before and after the payload release, and \emph{settling time} denotes the duration required for the error to converge and remain within a $0.05$\,m band. Lower values in both indicate greater robustness and faster adaptation.
}
\label{tracking_plots}

\end{figure*}

\begin{table*}[b]
\footnotesize
\renewcommand{\arraystretch}{1.3}
\caption{\small Performance comparison under different swing payloads and velocities (RMSE in meters)}
\centering
\scalebox{0.98}{
\begin{tabular}{lccc ccc ccc}
\toprule
\multirow{2}{*}{\textbf{Method}} & \multicolumn{3}{c}{\textbf{Payload = 75 g (38\% throttle)}} & \multicolumn{3}{c}{\textbf{Payload = 125 g (46\% throttle)}} & \multicolumn{3}{c}{\textbf{Payload = 175 g (55\% throttle)}} \\
\cmidrule(r){2-4} \cmidrule(r){5-7} \cmidrule(r){8-10}
 & $0.5$ m/s & $1.5$ m/s & $2.5$ m/s & $0.5$ m/s & $1.5$ m/s & $2.5$ m/s & $0.5$ m/s & $1.5$ m/s & $2.5$ m/s \\
\midrule
\textbf{Nominal MPC}        & 0.086 & 0.089 & 0.092 & 0.096 & 0.102 & 0.115 & 0.138 & 0.183 & 0.269 \\
\textbf{$\mathcal{L}_1$-adaptive}             & 0.081 & 0.082 & 0.085 & 0.085 & 0.089 & 0.096 & 0.118 & 0.152 & 0.218 \\
\textbf{Proto-MPC}             & 0.080 & 0.081 & 0.083 & 0.084 & 0.087 & 0.094 & 0.109 & 0.138 & 0.198 \\ 
\textbf{GP-MPC}          & 0.077 & 0.079 & 0.081 & 0.081 & 0.084 & 0.091 & 0.103 & 0.129 & 0.187 \\
\textbf{MLP-MPC}          & 0.077 & 0.078 & 0.080 & 0.079 & 0.082 & 0.088 & 0.101 & 0.125 & 0.178 \\
\textbf{Ours (w/o CP)}     & {0.065} & {0.066} & {0.067} & {0.068} & {0.070} & {0.074} & {0.080} & {0.097} & {0.135} \\
\textbf{Ours}& \textbf{0.054} & \textbf{0.054} & \textbf{0.055} & \textbf{0.056} & \textbf{0.058} & \textbf{0.062} & \textbf{0.071} & \textbf{0.086} & \textbf{0.120} \\
\bottomrule
\end{tabular}
}
\label{tab:adaption_payload_velocity}
\vspace{-2mm}
\end{table*}
Performance across payload/velocity settings is summarized in Table~\ref{tab:adaption_payload_velocity}. Figure~\ref{tracking_plots} illustrates one representative, challenging setting ($175$\,g at $2.5$\,m/s): the left column shows $x$- , $y$- and $z$-axis tracking errors over time respectively. The bar plots on the right, aggregate metrics \emph{across all configurations} (10 trials each). We report RMS tracking error (combined and $z$-only) and two drop-specific metrics computed on the $z$-axis: 
\begin{itemize}
    \item \emph{shift magnitude} $\triangle e_z := \big|e_z(t_\mathrm{drop}^+) - e_z(t_\mathrm{drop}^-)\big|$, the instantaneous change at payload release; 
    \item \emph{settling time}, the time from $t_\mathrm{drop}$ until $|e_z(t)| \le 0.05$\,m thereafter.
\end{itemize}

In the time series, $x$/$y$ errors are broadly similar across methods during quasi-steady segments, with distinctions emerging at large dips and rises where accelerations and swing-induced lateral forces peak; the ordering at these extrema mirrors the aggregated trends in Table~\ref{tab:adaption_payload_velocity}. The $z$-axis makes the contrast explicit: prior to the drop, methods follow the same ranking as in the table; at the drop ($\sim\!10$\,s), all controllers see a positive step due to the thrust-to-weight jump, after which our changepoint-aware model adapts fastest and stabilizes at the lowest steady error. The no-changepoint variant adapts more slowly, while nominal MPC and $\mathcal{L}_1$ exhibit prolonged transients; SysID fails to re-stabilize reliably.

The differences in recovery behavior stem from the adaptation mechanism of each method. Nominal MPC uses a fixed model that neither adapts to regime changes nor accounts for continuous, state-dependent disturbances from the swinging payload, leading to bias even before the drop and sustained $z$-error afterwards. $\mathcal{L}_1$-adaptive control assumes matched uncertainties. The sudden, nonlinear shift at payload release is matched, but if the filter bandwidth is not chosen appropriately, the abrupt change can lead to overshoot and long settling \cite{1657243, hovakimyan2010}. Proto-MPC adapts only within a convex hull of pre-trained prototypes, limiting its ability to capture complex, time-varying payload disturbances. GP-MPC can represent such effects in principle, but does not incorporate predictive uncertainty into the MPC, preventing gain modulation that would enhance robustness when model confidence is low; combined with slow adaptation under sparse online samples~\cite{park2020gaussian}, this results in persistent post-drop errors. MLP-MPC benefits from gain modulation via a UKF~\cite{wan2000unscented}, aiding lateral tracking, but its last-layer SGD updates are brittle~\cite{saviolo2023active, jastrzkebski2017three, smith2017cyclical, song2020rapidly, kaushik2020fast}, yielding inconsistent adaptation and slower $z$-error recovery. Our no-changepoint variant maintains good steady-state tracking, but retains stale parameters after the drop, delaying convergence. In contrast, our changepoint-aware update resets posterior uncertainty upon detecting the drop, enabling rapid re-identification of the altered dynamics and minimizing both $\triangle e_z$ and settling time.

Across all payload/velocity combinations, the aggregated bar plots show that our method achieves the lowest RMS errors and the smallest $\triangle e_z$ with the shortest settling times, with consistent improvements over nominal MPC (e.g., $37\%$ at $75$\,g, $0.5$\,m/s and $55\%$ at $175$\,g, $2.5$\,m/s). These results align with the open-loop validation: the learned representation remains well-conditioned and informative on real flights, and the changepoint-aware updates enable rapid post-drop re-calibration in closed loop.

\section{Conclusion}
\label{sec:conclusion}
We introduced a structured latent-variable framework for online control under nonstationary dynamics, combining Bayesian linear regression with changepoint-aware adaptation. By decoupling offline representation learning from lightweight online updates, our method enables rapid and robust tracking across dynamic conditions. The incorporation of predictive uncertainty into the control cost further enhances resilience to model uncertainty and perception noise. Theoretically, we established adaptive regret bounds under piecewise-stationary regimes with unobserved shifts. Empirically, our approach demonstrates strong performance in both simulation and real-world quadcopter experiments involving swinging payloads and mid-flight mass drops. These results underscore the importance of modeling and responding to dynamic shifts in real time, and highlight the promise of structured Bayesian adaptation for reliable, uncertainty-aware control in real-world robotics.
\vspace{-10pt}
\section{Limitations}
Despite strong empirical performance, our framework is based on several assumptions and design choices that may pose some limitations and warrant further investigation.  First, by decoupling offline representation learning (via a VLD) from online adaptation of a linear decoder, we rely on a fixed latent embedding that may drift under prolonged deployment; exploring ``slow update'' strategies for the encoder could mitigate representation collapse and maintain well‐conditioned Bayesian updates. Second, while our adaptive regret bounds assume piecewise-stationary dynamics with a bounded number of changepoints, this theoretical assumption does not limit the algorithm’s practical ability to handle continuous or rapid shifts; future work should seek to relax these assumptions to derive tighter guarantees under more general nonstationarity. Third, although our beam‐size ablation on the quadrotor shows inference latency remains around 7.6 ms across beam sizes up to 30, thanks to JAX/XLA vectorization~(see Appendix ~\ref{app:ablation}), deploying on resource‐constrained embedded hardware or very high-dimensional robotic platforms may pose challenges, motivating investigation of even more scalable update schemes. Finally, integrating a systematic study of uncertainty calibration under model mismatch represents an exciting opportunity to strengthen confidence and reliability in extreme or unmodeled regimes and inform future algorithmic enhancements.

\appendices

\section{Proofs}
\subsection{Proof of Lemma \ref{lemma:posterior_consistency}}\label{app:posterior_consistency}
\begin{proof}
For a single output dimension $j$, the posterior at time $T$ is defined as~\cite{ghosal2017fundamentals, murphy2023probabilistic}:
\begin{align*}
    \Sigma^j_T &= \left( (\Sigma_0^j)^{-1} + \frac{1}{\sigma_j^2} Z_T^\top Z_T \right)^{-1}, \\
    \mu^j_T &= \Sigma^j_T \left( (\Sigma_0^j)^{-1} \mu_0^j + \frac{1}{\sigma_j^2} Z_T^\top \delta^j_{1:T} \right),
\end{align*}
where \( Z_T = [z_1^\top, \ldots, z_T^\top] \) is the design matrix, and \( \delta^j_{1:T} \) is the vector of targets for dimension \( j \).
Now, $\displaystyle{A_T^j\coloneqq (\Sigma_0^j)^{-1}+\frac{1}{\sigma^2_j}Z_T^\top Z_T\succ 0}$; $\enskip \Sigma^j_T=(A^j_T)^{-1}$.\\
By the Loewner-order monotonicty of inversion, if $A_T\succeq B_T\succ 0$, then $A_T^{-1}\preceq B_T^{-1}$~\cite[Corollary~$7.7.4$]{matrixAnalysis}. Now since $(\Sigma_0^j)^{-1}\succeq 0$, we have:
\[
    \norm{\Sigma^j_T}_2 \leq \frac{\sigma_j^2}{\lambda_{\min}(Z_T^\top Z_T)}.
\]
By the condition of design-growth, the denominator diverges, so $\|\Sigma^j_T\|_2\to 0$ and therefore $\operatorname{tr}\fbr{\Sigma^j_T}\to 0$.
Using the bias–variance decomposition:
\[
\mathbb{E}\left[\|\mu^j_T - \theta^{\star j}\|^2\right] = \operatorname{tr}(\Sigma^j_T) + \|\mathbb{E}[\mu^j_T] - \theta^{\star j}\|^2.
\]

\noindent Now, the OLS estimate converges to the true parameter \( \theta^{\star j} \) in expectation under standard conditions. Hence, both terms vanish, completing the proof.
\end{proof}

\subsection{Proof of Lemma \ref{lemma:bounded_variance}}
\label{app:bounded_variance}
\begin{proof}
    Let $k_s$ denote the the first index of the current stationary segment $s$ (i.e., one step after the most recent changepoint), and write $S_s={k_s, \dots k}$ for the indices observed so far in this segment. From \eqref{eqn:posterior_covariance}, the posterior covariance update with tempering is given by
    \[ \fbr{\Sigma^j_{k,h}}^{-1}=\gamma_k\fbr{\Sigma^j_{k-1,h}}^{-1}+\tfrac{1}{\sigma_j^2}z_{k-1}z_{k-1}^\top.\]
    Unrolling this within the segment and using $\gamma_t\ge \beta^2>0$ yields the lower bound on precision
    \[\fbr{\Sigma^j_{k,h}}^{-1}\succeq \tfrac{1}{\sigma_j^2}\sum_{t\in S_s}z_tz_t^\top \implies \Sigma^j_{k,h} \preceq \fbr{\tfrac{1}{\sigma_j^2}\sum_{t\in S_s}z_tz_t^\top}^{-1} \]
    By the Rayleigh quotient inequality~\cite{matrixAnalysis} followed by the minimum eigenvalue assumption inside the segment, we have
    \begin{align*}
        z_k^\top\Sigma^j_{k,h}z_k & \le \norm{z_k}^2\lambda_{\max}\left(\left(\tfrac{1}{\sigma_j^2}\sum_{t\in S_s}z_tz_t^\top\right)^{-1}\right) \\
        & \le \frac{R^2\sigma_j^2}{\lambda_{\min}(\sum_{t\in S_s}z_tz_t^\top)} \le \frac{R^2\sigma_j^2}{\alpha\, t(k)}.
    \end{align*}
    Immediately after a changepoint, before any data in the new segment is incorporated, the tempered prior is no more diffuse than the offline prior $\tau^2 I$ (one may equivalently inject the offline prior precision $\tau^{-2}I$ at changepoints), hence
    \[z_k^\top\Sigma^j_{k,h}z_k \le \tau^2R^2 \text{ at } t(k)=1.\]
    Combining the two bounds gives
    \[z_k^\top\Sigma^j_{k,h}z_k \le \min\left\{\tau^2R^2, \frac{R^2\sigma_j^2}{\alpha\, t(k)}\right\}\leq \tau^2R^2+\frac{R^2\sigma_j^2}{\alpha},\]
uniformly in $k$ and independently of $\kappa$ in the instantaneous sense. Moreover, when viewing the segmentwise worst-case envelope across at most $\kappa$ resets, the reset cost $\tau^2R^2$ can be incurred at most $\kappa$ times, yielding at most linear dependence on $\kappa$ when aggregated.
\end{proof}

\subsection{Changepoint Scoring}
\label{app:changepoint_scoring}
We begin by deriving the recursive expression of the hypothesis scores. Each hypothesis $h$ corresponds to a sequence of changepoint decisions $c_{1:k}^{(h)}$, and its associated score is defined as the joint log-probability:
\[
\mathcal{L}_{k,h} := \log p(\delta_{1:k}, c_{1:k}^{(h)} \mid x_{1:k}, u_{1:k})
\]

\noindent By applying the chain rule of probability and taking logs:
\begin{multline*}
    \log p(\delta_{1:k}, c_{1:k} \mid x_{1:k}, u_{1:k}) \\
    = \log p(\delta_{1:k-1}, c_{1:k-1} \mid x_{1:k-1}, u_{1:k-1})\\ + \log p(c_k \mid c_{1:k-1}, x_{1:k}, u_{1:k}) + \log p(\delta_k \mid c_{1:k}, x_{1:k}, u_{1:k})
\end{multline*}

To simplify the predictive term, we note that under our Bayesian linear model,  \( \delta_k = x_{k+1} - f_{\rm nom}(x_k, u_k) \) depends only on the current changepoint decision \( c_k \), the latent input \( z_k = \phi(x_k, u_k) \), and the decoder posterior from hypothesis \( h \). Thus, $p(\delta_k \mid c_{1:k}, x_{1:k}, u_{1:k}) = p(\delta_k \mid z_k, c_k)$. \\Substituting back, we obtain the final recursive expression:
\[
\mathcal{L}_{k,h} = \mathcal{L}_{k-1,h} + \log p(\delta_k \mid z_k, c_k) + \log p(c_k \mid c_{1:k-1}, x_{1:k}, u_{1:k}).
\]

This expression enables efficient online computation of hypothesis scores without re-evaluating the full history at each step.

We now proceed to derive the marginal likelihood (model evidence) of the observation $\delta_k$ under the Bayesian linear model with Gaussian prior and likelihood. For each output dimension $j = 1, \dots, d$, we have:
\begin{align*}
    \theta^j &\sim \mathcal{N}(\mu^j_{k-1,h}, \Sigma^j_{k-1,h} / \gamma_k), \\
    \delta_k^j \mid \theta^j &\sim \mathcal{N}(z_k^\top \theta^j, \sigma_j^2),
\end{align*}
where $\gamma_k = 1$ if $c_k = 0$, and $\gamma_k = \beta^2$ if $c_k = 1$.

The marginal likelihood is obtained by integrating out the decoder weights $\theta^j$:
\[
p(\delta_k^j \mid z_k, c_k) = \int p(\delta_k^j \mid \theta^j, z_k) \, p(\theta^j \mid c_k) \, d\theta^j
\]

\noindent This integral is analytically tractable due to Gaussian conjugacy:
\[
p(\delta_k^j \mid z_k, c_k) = \mathcal{N}\left(\delta_k^j \mid z_k^\top \mu^j_{k-1,h},\ \frac{1}{\gamma_k} z_k^\top \Sigma^j_{k-1,h} z_k + \sigma_j^2\right)
\]

\noindent Thus, the log marginal likelihood for output $j$ becomes:
\begin{multline*}
    \log p(\delta_k^j \mid z_k, c_k) = -\frac{1}{2} \log \left(2\pi \left( \tfrac{1}{\gamma_k} z_k^\top \Sigma^j_{k-1,h} z_k + \sigma_j^2 \right) \right) \\
    - \frac{1}{2} \frac{\left( \delta_k^j - z_k^\top \mu^j_{k-1,h} \right)^2}{ \tfrac{1}{\gamma_k} z_k^\top \Sigma^j_{k-1,h} z_k + \sigma_j^2 }
\end{multline*}

\noindent Summing over all $j$ gives us the total log marginal likelihood:
\begin{multline*}
    \log p(\delta_k \mid z_k, c_k) = -\frac{1}{2} \sum_{j=1}^d \Bigg[ \log (2\pi( \tfrac{1}{\gamma_k} z_k^\top \Sigma^j_{k-1,h} z_k + \sigma_j^2 )) \\
    + \frac{\left( \delta_k^j - z_k^\top \mu^j_{k-1,h} \right)^2}{ \tfrac{1}{\gamma_k} z_k^\top \Sigma^j_{k-1,h} z_k + \sigma_j^2 } \Bigg]
\end{multline*}

\subsection{Proof of Theorem \ref{thm:regret}}
\label{proof:regret}
\begin{proof}
Let $h^\star$ denote the hypothesis corresponding to the true changepoint sequence (i.e., the optimal piecewise stationary model). Let $\mathcal{H}_k$ denote the beam of top-$K$ hypotheses maintained at time $k$ by the algorithm. Then, the cumulative regret decomposes as:
\begin{multline*}
    \underbrace{\sum_{k=1}^T \left( \mathbb{E}\left[\|\delta_k - \hat{\delta}_k\|^2\right] - \mathbb{E}\left[\|\delta_k - \hat{\delta}_k^{(h^\star)}\|^2\right] \right)}_{\text{(A) Beam approximation error}} \\
    + \underbrace{\sum_{k=1}^T \left( \mathbb{E}\left[\|\delta_k - \hat{\delta}_k^{(h^\star)}\|^2\right] - \|\delta_k - \theta_{[k]}^\star z_k\|^2 \right)}_{\text{(B) Estimation error within segments}}.
\end{multline*}

\textbf{(A) Beam approximation error.}  
The algorithm performs online aggregation (e.g., exponential weighting) over the $K$ hypotheses in the beam $\mathcal{H}_k$. Since the squared loss is exponentially concave, the regret compared to the best hypothesis in the beam is at most $\mathcal{O}(\log K)$ over $T$ rounds~\cite{cesa2006prediction}. Therefore, the cumulative beam approximation error is $\mathcal{O}(\log K)$. Since $K$ is typically small and fixed in practice, this term is negligible compared to the dominant $\log T$ dependence.

\textbf{(B) Estimation error within each stationary segment.}  
Within each stationary segment, the problem reduces to standard online learning under a fixed target. The cumulative loss over a stationary segment becomes strongly convex for bounded and non-degenerate $z_k$. Thus the cumulative estimation error within a segment of length $T_j$ is at most $\mathcal{O}(\ell \log T_j)$. Summing over all $\kappa+1$ stationary segments and using Jensen's inequality, we obtain:
\[
\sum_{j=1}^{\kappa+1} \mathcal{O}(\ell \log T_j) \leq \mathcal{O}(\kappa \ell \log T),
\]
where we used that $\sum_{j=1}^{\kappa+1} T_j = T$. An additional additive term of order $\log K!$ could arise from selecting among changepoint sequences in the beam search, but this is negligible compared to the leading terms.
Combining the two parts, the total regret satisfies:
\[
\mathcal{O}(\log K) + \mathcal{O}(\kappa \ell \log T) = \mathcal{O}(\log T + \kappa \ell \log T),
\]
where the $\log K$ term is dominated asymptotically by the $\log T$ term. This completes the proof.
\end{proof}

\begin{remark}
    The variance bound in Lemma \ref{lemma:bounded_variance} and the regret bound in Theorem \ref{thm:regret} are finite-time results that hold uniformly for all $T$, they require no asymptotic assumptions beyond the design conditions assumed in Lemma \ref{lemma:posterior_consistency}.
\end{remark}

\section{Discussions}
\subsection{Latent Representations and Posterior Conditioning}\label{appendix:posterior_conditioning}
\begin{figure*}[h]
    \centering
    \subfloat[Embedding Visualization.\label{fig:embedding}]{\includegraphics[width=0.49\linewidth]{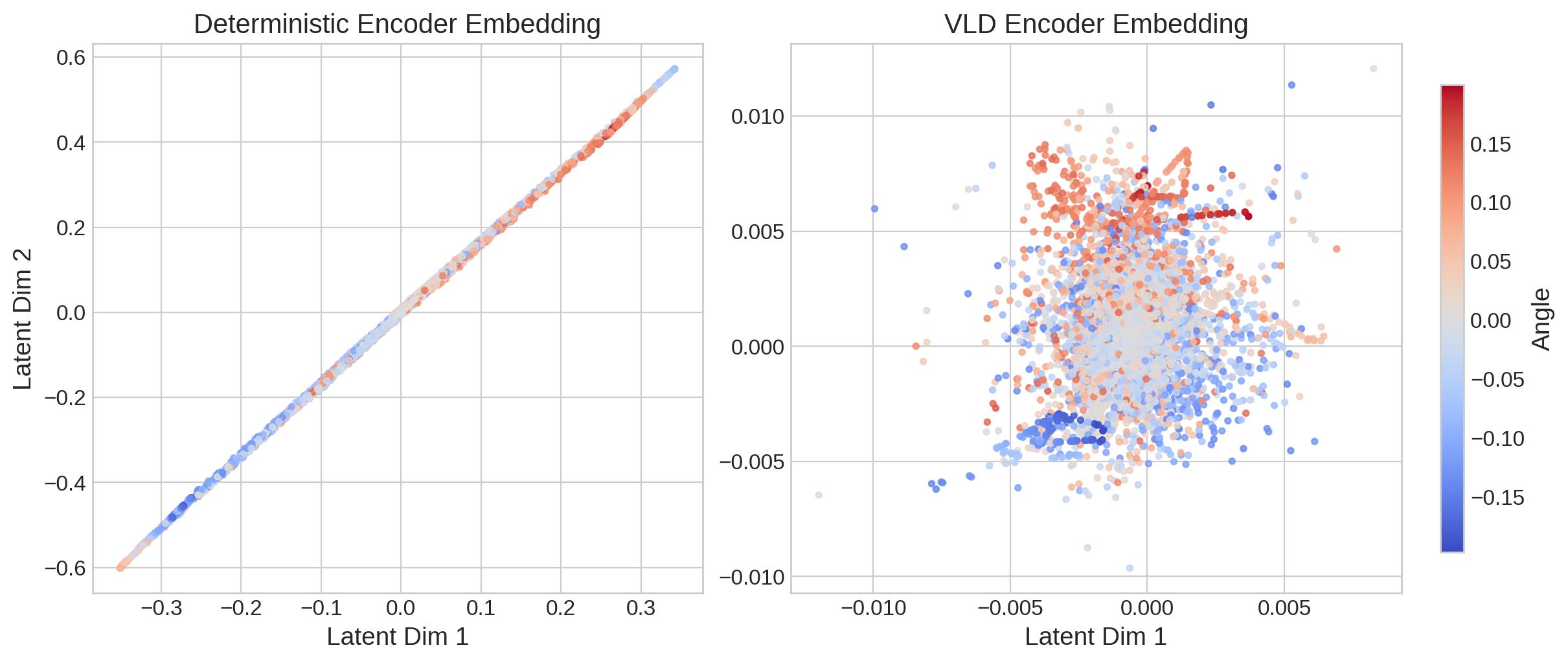}}
    \hfill
    \subfloat[Posterior Uncertainty Ellipses.\label{fig:posterior_ellipses}]{\includegraphics[width=0.49\linewidth]{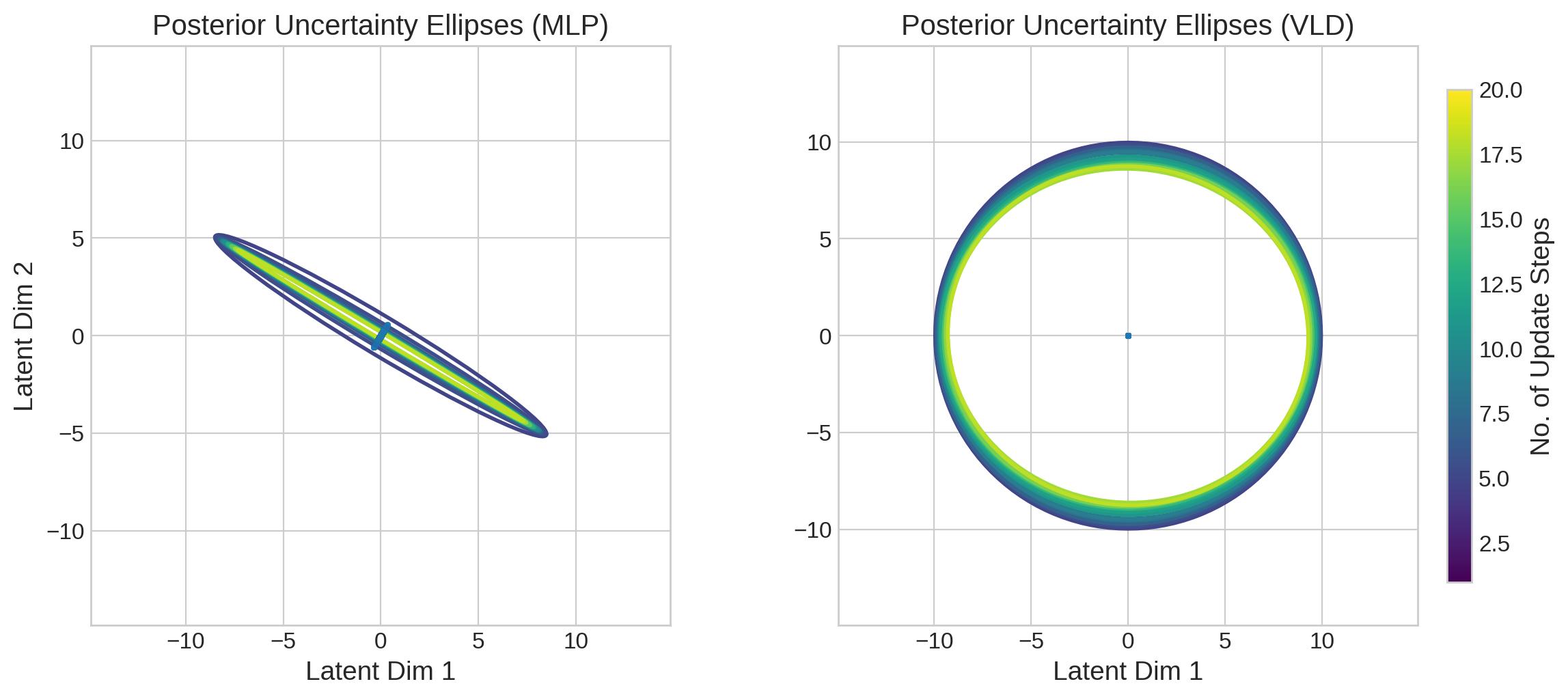}}
    \caption{Latent space and posterior behavior for deterministic and variational encoders, trained on offline data collected from a cartpole environment. Each model learns a 2D latent space and shares the same architecture and linear decoder, differing only in the encoder's formulation (deterministic vs.~variational). 
    \textit{Left:} The deterministic encoder collapses to a degenerate 1D manifold, encoding task-relevant information along a single principal direction. \textit{Right:} The VLD encoder yields well-distributed, isotropic latent representations that produce stable, well-conditioned posterior updates under Bayesian linear regression.}
    \label{fig:vae_vs_mlp}
\end{figure*} 
To obtain initial parameters $\fbr{\phi, \theta_0}$ from offline training data, a natural objective is the maximum a posteriori (MAP) of the form:
\begin{multline*}
    \mathcal{L}_{\text{MAP}}(\phi, \theta_0) = \sum_{k=1}^N \fbr{x_{k+1} - \fbr{f_{\rm nom}(x_k,u_k)+\theta_0 \phi(x_k, u_k)}}^2 \notag\\
    + \lambda_\theta \norm{\theta_0}^2 + \lambda_\phi \norm{\phi}^2,
\end{multline*}
which corresponds to ridge-regularized regression in latent space. While simple, this formulation often yields degenerate features and poorly conditioned decoders, undermining the stability of Bayesian updates during deployment. To better understand the impact of encoder design on downstream adaptation, we visualize and analyze the learned latent features and their corresponding posterior dynamics. Figure \ref{fig:vae_vs_mlp} compares the behavior of a deterministic MLP encoder and a Variational Latent Dynamics (VLD) model, both trained on offline data collected from an inverted pendulum environment. Each model maps inputs $(x_k, u_k)$ into a shared two-dimensional latent space, followed by a fixed linear decoder.

\paragraph{Latent Embedding Structure (Fig.~\ref{fig:embedding}).}  
We visualize the latent codes produced by each encoder, with each point corresponding to an input $(x_k, u_k)$ and colored by the angle of the pendulum. While angle is directly observed and part of the input state, it is a physically meaningful quantity that governs the system dynamics and provides a natural diagnostic for the structure of the learned representation. The deterministic encoder exhibits a collapsed geometry, with most points aligned along a single direction, suggesting a failure to utilize the full latent capacity. In contrast, the VLD encoder yields a smoother, more isotropic embedding in which the angle is distributed more uniformly across the latent space. This structure is more amenable to downstream adaptation and uncertainty estimation, especially under covariate shift.

\paragraph{Posterior Uncertainty Dynamics (Fig.~\ref{fig:posterior_ellipses}).}  
We simulate Bayesian adaptation by incrementally updating the linear decoder via Bayesian Linear Regression (BLR). At each step, we condition the posterior on an increasingly large subset of the offline data and visualize the resulting posterior covariance as a one-standard-deviation ellipse at a fixed latent input. The deterministic encoder produces highly anisotropic and unstable posterior shapes (ellipses are elongated and misaligned, reflecting high uncertainty in collapsed directions), reflecting poor conditioning in the latent basis. The VLD encoder, on the contrary, yields isotropic and stable posterior contours that shrink smoothly with more observations, indicating a well-structured latent geometry for adaptation.

\subsection{Statistical Interpretation of Changepoint Detection as a Likelihood Ratio Test}
\label{app:likelihood-ratio_discussion}
In this section, we attempt to formalize the connection between our changepoint posterior computation and classical statistical hypothesis testing, particularly the likelihood ratio test (LRT). The posterior update can be interpreted as performing a Bayesian model comparison between two competing hypotheses at each time step $k$:
\begin{itemize}
    \item $\mathcal{H}_0$: No changepoint occurred at time $k$; the decoder parameters $\theta$ follow the untempered posterior from time $k-1$.
    \item $\mathcal{H}_1$: A changepoint occurred at time $k$; the prior is a tempered version of the posterior from time $k-1$.
\end{itemize}

Let $p(\delta_k \mid z_k, \mathcal{H}_0)$ and $p(\delta_k \mid z_k, \mathcal{H}_1)$ denote the predictive marginal likelihoods under each hypothesis, computed via marginalization over the Gaussian posterior or tempered prior as detailed in the main text. Then, the posterior probability of a changepoint given current data is:

\begin{equation*}
    p(c_k = 1 \mid \cdot) = \frac{\pi \cdot p(\delta_k \mid z_k, \mathcal{H}_1)}{\pi \cdot p(\delta_k \mid z_k, \mathcal{H}_1) + (1 - \pi) \cdot p(\delta_k \mid z_k, \mathcal{H}_0)},
\end{equation*}

where $\pi \in (0, 1)$ is the prior changepoint probability. This expression arises from Bayesian model averaging, and is equivalent to computing the posterior model probability under a finite mixture of models:
\[
p(\delta_k \mid z_k) = \pi \cdot p(\delta_k \mid z_k, \mathcal{H}_1) + (1 - \pi) \cdot p(\delta_k \mid z_k, \mathcal{H}_0).
\]

The posterior $p(c_k = 1 \mid \cdot)$ can be seen as a \textit{soft generalization of the likelihood ratio test (LRT)}, where instead of choosing the most likely hypothesis outright (as in classical LRTs), we compute the probability that the changepoint hypothesis better explains the data, given a Bayesian prior. Formally, the \textit{likelihood ratio} is given by:
\begin{equation*}
    \Lambda_k = \frac{p(\delta_k \mid z_k, \mathcal{H}_1)}{p(\delta_k \mid z_k, \mathcal{H}_0)}.
\end{equation*}
The changepoint posterior becomes:
\begin{equation*}
    p(c_k = 1 \mid \cdot) = \frac{\pi \cdot \Lambda_k}{\pi \cdot \Lambda_k + (1 - \pi)}.
\end{equation*}

This ratio acts as a Bayesian test statistic, and the resulting posterior resembles a sigmoid transformation of the log-likelihood ratio, scaled by the prior odds:
\begin{equation*}
    \log \frac{p(c_k = 1 \mid \cdot)}{p(c_k = 0 \mid \cdot)} = \log \Lambda_k + \log \frac{\pi}{1 - \pi}.
\end{equation*}
This formulation is closely related to the framework of Bayesian online changepoint detection (BOCPD) introduced by \cite{adams2007bayesian}, and further generalized in sequential multiple changepoint settings~\cite{fearnhead2007line}. In BOCPD, changepoint detection is treated as recursive Bayesian inference over run-lengths, with model evidence guiding the posterior over hypotheses. The posterior update can be viewed as a structured variant of this idea, where we track multiple changepoint hypotheses using beam search and perform soft inference via marginal likelihood ratios at each step.

The use of tempered priors is also justifiable. Given a posterior $p(\theta \mid \mathcal{D}_{1:k-1})$, raising the density to a power $\beta \in (0, 1)$ and renormalizing yields a tempered prior:
\[
\tilde{p}_\beta(\theta) = \frac{p(\theta \mid \mathcal{D}_{1:k-1})^\beta}{Z_\beta},
\]
where $Z_\beta$ is the normalizing constant. This construction is known as a power prior~\cite{ibrahim2000power} and is widely used in Bayesian updating under partial trust or model mismatch. In our setting, this allows the model to softly ``forget'' past information while retaining its structure.

\subsection{Practical Considerations for Hyperparameter Selection}\label{app:hyperparam_selection}
The hyperparameters $\pi$, $\beta$, $K$, $\tau^2$, and $\sigma^2$ are problem-dependent and must be set in accordance with the operating environment. We note a few general considerations for applying our method:
\begin{itemize}
    \item Changepoint prior $\pi$: Determines the prior probability assigned to a changepoint at each step. A larger value makes the algorithm more responsive to potential regime shits, but increases the likelihood of spurious resets. A smaller value reduces false alarms at the cost of slower adaptation. In practice, values in the range $[0.01, 0.10]$ should provide a reasonable trade-off.
    \item Prior tempering $\beta$: Determines the strength of the reset at a detected changepoint. Smaller values enforce more aggressive forgetting and faster recovery, whereas values closer to one retain more information and yield more conservative adaptation. In practice, $\beta$ can be selected from the range $[0.90, 0.999]$.
    \item Beam size $K$: Controls the number of changepoint hypotheses maintained. Larger values increase robustness in ambiguous scenarios, but may also increase computational cost. In practice, it should be large enough that the prediction accuracy no longer improves appreciably, subject to the available computational budget.
    \item Decoder prior variance $\tau^2$: Sets the degree of confidence in the offline-trained parameters. Smaller values place more weight on the prior and are appropriate when the offline model is of high quality. Larger values allow for faster adaptation, which can be beneficial if the offline model is less reliable. In practice, $\tau^2$ can be selected from the range $[10^{-3}, 10^{-1}]$.
    \item Observation noise variance $\sigma^2$: Scales the likelihood model. Larger values lead to smoother, more conservative updates, while smaller values make the posterior more sensitive to noise. In practice, $\sigma^2$ can be selected from the range $[0.01, 0.10]$.
    
\end{itemize}

\section{Experiment and Implementation Details}

\subsection{Cartpole Experiments}
\label{app:cartpole}
\subsubsection{Dynamic Model}
The true (real-world) cart-pole system dynamics incorporate non-ideal effects such as friction and external disturbances. The continuous-time equations of motion are given by:
{\footnotesize
\begin{align*}
\ddot{x} &= \frac{u + u_{\text{dist}} - \mu_c \dot{x} + m \sin(\theta)\left(L \dot{\theta}^2 + g \cos(\theta)\right)}{M + m \sin^2(\theta)} \\
\ddot{\theta} &= \frac{-(u + u_{\text{dist}})\cos(\theta) - \mu_p \dot{\theta} - m L \dot{\theta}^2 \sin(\theta) \cos(\theta) + (M + m) g \sin(\theta)}{L(M + m \sin^2(\theta))}
\end{align*}
}

Here, \( x \) denotes the horizontal position of the cart, \( \theta \) is the pole angle measured from the upright vertical position, and \( u \in \mathbb{R} \) is the control force applied to the cart. The nominal physical parameters used are:
$M = 1.0, \quad m = 0.1, \quad L = 1.5, \quad g = 9.81$.
The non-ideal terms include friction and an additive disturbance modeled as:
$
\mu_c = 0.25 \quad \text{(cart friction)}, \quad 
\mu_p = 0.05 \quad \text{(pole friction)}, \quad 
u_{\text{dist}} \sim \mathcal{N}(-0.5, 0.5) \quad \text{(external disturbance)}$. For data collection, a nominal MPC controller is employed. Initial states are sampled randomly from a predefined distribution. The control objective is to stabilize the system at the upright equilibrium, defined by the reference state:
$ x_{\text{ref}} = \begin{bmatrix} 0 & 0 & 0 & 0 \end{bmatrix}^\top$.

The nominal dynamics used in the model predictive controller are based on simplified physics, omitting friction and disturbances. The equations are discretized using forward Euler integration:
{
\begin{align*}
\ddot{x} &= \frac{u + m \sin(\theta)\left(L \dot{\theta}^2 + g \cos(\theta)\right)}{M + m \sin^2(\theta)} \\
\ddot{\theta} &= \frac{-u \cos(\theta) - m L \dot{\theta}^2 \sin(\theta) \cos(\theta) + (M + m) g \sin(\theta)}{L(M + m \sin^2(\theta))}
\end{align*}
}
with nominal parameters: $
M = 1.7, \quad m = 0.25, \quad L = 1.7, \quad g = 9.81$. In this nominal model, system parameters are only approximately known and deviate from the true values, which leads to model uncertainties. Furthermore, the omission of friction and disturbances introduces a structured mismatch between the dynamics used for control and the actual system behavior.

\subsubsection{Training and Inference}
Table~\ref{tab:training_inference_configs} summarizes the configurations used during offline training and online inference for the cartpole experiments.

\begin{table}[h]
\centering
\footnotesize
\renewcommand{\arraystretch}{1.3}
\caption{Configurations for Cartpole.}
\label{tab:training_inference_cartpole}
\begin{minipage}[t]{0.48\textwidth}
\centering
\textbf{(a) Offline Training} \\
\vspace{0.5em}
\begin{tabular}{l|l}
\toprule
\textbf{Component} & \textbf{Description} \\
\midrule
Encoder Architecture ($\phi$) & 3-layer MLP: [8, 16, 8] \\
Activation & ELU \\
Latent Dimension ($\ell$) & 2 \\
KL Weight ($\lambda_{\text{KL}}$) & 0.1 \\
Optimizer & Adam  \\
Learning Rate & $1 \times 10^{-3}$ \\
Batch Size & 128 \\
Epochs & 50 \\
\bottomrule
\end{tabular}
\end{minipage}
\hfill
\vspace{0.8ex}
\begin{minipage}[t]{0.48\textwidth}
\centering
\textbf{(b) Online Inference} \\
\vspace{0.5em}
\begin{tabular}{l|c}
\toprule
\textbf{Parameter} & \textbf{Value} \\
\midrule
Beam size \( (K) \) & 5 \\
Changepoint prior \( (\pi) \) & 0.05 \\
Prior tempering \( (\beta) \) & 0.9 \\
Decoder variance \( (\tau^2) \) & $0.1$ \\
Obs. noise \( (\sigma^2_j) \) & $0.1$ \\
\bottomrule
\end{tabular}
\end{minipage}
\end{table}

\subsection{Quadrotor Experiments} \label{app:quad_details}
\subsubsection{Quadrotor Nominal Dynamics}
The quadrotor is modeled as a rigid body with six degrees of freedom, governed by the continuous-time dynamics:
\begin{align}
\dot{p} &= v,\quad 
\dot{v} = \frac{1}{m} f z_B + g,\nonumber \\ 
\dot{q} &= \frac{1}{2} q \otimes \begin{bmatrix} 0 \\ \omega \end{bmatrix},\quad 
\dot{\omega} = J^{-1}(M - \omega \times J\omega),
\end{align}
where \(p, v \in \mathbb{R}^3\) denote position and velocity in the inertial frame, \(q \in \mathbb{S}^3\) is the unit quaternion for orientation, and \(\omega \in \mathbb{R}^3\) is the angular velocity in the body frame. The control input \(u = [f z_B, M^\top]^\top\) comprises the total thrust \(f z_B \in \mathbb{R}\) and body torques \(M \in \mathbb{R}^3\) and \( g \) denotes the gravitational acceleration. To incorporate these dynamics into the MPC framework, we discretize them using a Runge-Kutta integrator, resulting in the discrete-time system:
$
x_{k+1} = f_{\text{nom}}(x_k, u_k)$, where $x_k = [p_k^\top, v_k^\top, q_k^\top, \omega_k^\top]^\top \in \mathbb{R}^{13}$ and $u_k \in \mathbb{R}^4$.

\subsubsection{Quadrotor Hardware Setup} 
The experimental platform consists of a 10-camera OptiTrack motion capture system and a quadrotor with an approximate mass of 0.55\,kg. The quadrotor is built on an customized 152 mm frame and is equipped with T-Motor F1404 3800KV\,KV brushless motors, 3-inch propellers, and powered by a 4S LiPo battery. The vehicle is controlled by an NxtPX4 flight controller running the custom PX4 firmware parameter and features an onboard Jetson Orin NX computer executing the \textit{MAVROS} package. A host computer, equipped with an NVIDIA GeForce RTX 4080 GPU, is used for inference tasks. Sensor data from the onboard computer is transmitted via WiFi to the host machine, where it is processed and control commands are generated. These commands, including the desired body rates and collective thrust computed by a model predictive controller (MPC), are transmitted back to the quadrotor at 30\,Hz and executed by an onboard PID controller \cite{saviolo2023active, djeumou2023learn}. This hardware configuration is representative of typical consumer-grade drone systems. The payload is suspended using a lightweight cable, with one end connected to an electromagnet. The electromagnet is controlled via Jetson GPIOs, allowing the payload to be released mid-flight by toggling power ON or OFF at desired waypoint. The electromagnet weighs approximately $20$ g.

\subsubsection{Data Collection}
An expert pilot flew the quadrotor along randomized trajectories with varying velocities and rigidly attached payloads (10, 100, and 200\,g), collecting 3 minutes of data per payload at 100\,Hz. The dataset consists of time-aligned state-control pairs,
$\mathcal{D} = \left[ (x(t_1), u(t_1)),\ (x(t_2), u(t_2)),\ \dots,\ (x(t_N), u(t_N)) \right]^\top,
$ sampled at timestamps \( \mathcal{T} = \{t_1, t_2, \dots, t_N\} \).

\subsubsection{Quadrotor MPC gains} 
To solve the OCP at each control step, we adopt a \textit{direct shooting} approach, where only the control inputs $\{u_{k+i}\}_{i=0}^{N-1}$ are treated as decision variables. The associated state trajectory is computed by forward simulating the system dynamics from the current state measurement. The resulting non-linear optimization problem is solved using \texttt{Optimistix}~\cite{optimistix}, which provides differentiable optimization routines in JAX~\cite{jax2018github}. Additional regularization terms for control smoothness (weight = 10) and soft constraint penalties (weight = 1000) are included in the cost function. This setup enables efficient, real-time optimization with full use of automatic differentiation and JIT compilation, without the need for external solver dependencies. Further details are provided in Table \ref{tab:mpc_hyperparams}. 
\begin{table}[h]
\centering
\footnotesize
\renewcommand{\arraystretch}{1.3}
\caption{Quadrotor MPC controller configuration.}
\begin{tabular}{l|c}
\toprule
\textbf{Parameter} & \textbf{Value} \\
\midrule
 \( Q \) & \( \mathrm{diag}([40, 40, 60, 10, 10, 10, 1, 10, 10, 10, 10, 10, 10]) \) \\
 \( P \) & \(  \mathrm{diag}([400, 400, 600, 20, 20, 20, 1, 10, 10, 10, 10, 10, 10]) \) \\
 \( R \) & \( \mathrm{diag}([10, 20, 20, 20]) \) \\
$U$ Limit & \( \text{min: } [0.1, -0.5, -0.5, -0.5], \text{max: } [20, 0.5, 0.5, 0.5] \) \\
 \( U_0 \) & Hover thrust + zero body moments \\
 N & $20$ (covering a $1$ second horizon)\\
 $\alpha_1$ & $10$ \\
 $\alpha_2$ & $10^3$\\
\bottomrule
\end{tabular}
\label{tab:mpc_hyperparams}
\end{table}

\subsubsection{Training and Inference}
Table~\ref{tab:training_inference_configs} summarizes the configurations used during offline training and online inference for the quadrotor experiments.
\begin{table}[h]
\centering
\footnotesize
\renewcommand{\arraystretch}{1.3}
\caption{Offline training and online inference configurations for Quadrotor.}
\label{tab:training_inference_configs}
\begin{minipage}[t]{0.48\textwidth}
\centering
\textbf{(a) Offline Training} \\
\vspace{0.5em}
\begin{tabular}{l|l}
\toprule
\textbf{Component} & \textbf{Description} \\
\midrule
Encoder Architecture ($\phi$) & 3-layer MLP: [64, 128, 64] \\
Activation & ELU \\
Latent Dimension ($\ell$) & 8 \\
KL Weight ($\lambda_{\text{KL}}$) & 0.1 \\
Optimizer & Adam  \\
Learning Rate & $1 \times 10^{-3}$ \\
Batch Size & 256 \\
Epochs & 100 \\
\bottomrule
\end{tabular}
\end{minipage}
\hfill
\vspace{0.8ex}
\begin{minipage}[t]{0.48\textwidth}
\centering
\textbf{(b) Online Inference} \\
\vspace{0.5em}
\begin{tabular}{l|c}
\toprule
\textbf{Parameter} & \textbf{Value} \\
\midrule
Beam size \( (K) \) & 15 \\
Changepoint prior \( (\pi) \) & 0.05 \\
Prior tempering \( (\beta) \) & 0.997 \\
Decoder variance \( (\tau^2) \) & $0.1$ \\
Obs. noise \( (\sigma^2_j) \) & $0.1$ \\
\bottomrule
\end{tabular}
\end{minipage}

\end{table}

\subsubsection{Baseline Implementations}
\label{app:baseline}
To evaluate the effectiveness of our proposed approach, we compare against several baselines for dynamics modeling and online adaptation. Where needed, we adapt training procedures and inference mechanisms to align with our evaluation scenarios involving varying payloads, target velocities, and abrupt disturbances.

\begin{enumerate}
\item \textbf{Nominal MPC:} A classical nonlinear MPC controller that uses only the known nominal dynamics model \( f_{\text{nom}} \), without any unmodeled dynamics. This baseline does not incorporate learning and serves to highlight the limitations of relying solely on nominal models.

\item \textbf{GP-MPC}~\cite{torrente2021data}: This baseline uses GPs to learn dynamics online. Following the original work, we use an ARD RBF kernel for per-output scalar GPs. The kernel hyperparameters of are selected by maximum likelihood optimization on the offline dataset. To maintain fairness, a FIFO buffer of capacity 50 is used to store the most recent state-control observations, and the GP is refit using these samples. The GP’s mean prediction is used to correct the nominal dynamics in the MPC rollout, but the predictive variance is not incorporated into the cost or constraints. While originally evaluated under aerodynamic disturbances, we adapt the original implementation to model the residuals induced by the swinging payload using an online update process.

\item \textbf{MLP-MPC}~\cite{saviolo2023active}: This baseline trains an MLP on offline data and fine-tunes the last-layer online using SGD. While the original work learns the full quadrotor dynamics offline using over \emph{58 minutes} of data, we slightly modify the approach to learn the residual dynamics instead. To ensure fairness, we use network architectures similar to our model: $[8, 16, 8, 8]$ for the Cartpole and $[64, 128, 64, 64]$ for the Quadrotor. During the online phase, the model is updated using the entire FIFO buffer with a learning rate of $1e-6$; a lower learning rate results in poor performance, while a higher one leads to instability and crashes. Following the original work, an Unscented Transformation (UT)~\cite{wan2000unscented} is applied to the MLP outputs to estimate epistemic uncertainty, which is then used to modulate the cost function in the MPC~\cite[Sec.III.D]{saviolo2023active}.

\item \textbf{Proto-MPC}~\cite{gu2024proto}: Proto-MPC is representative of a Meta-Learning baseline and uses an encoder-decoder architecture to model residual dynamics across task variations. During training, each trajectory is associated with a task-specific prototype, and the residual is predicted by a weighted average over decoders anchored to these prototypes. The original paper defines tasks based on wind conditions; in our case, we redefine tasks by rigidly attached payload mass (10, 100, 200\,g), and retrain the model accordingly. At test time, the swinging payload is treated as a novel condition, and residuals are inferred via attention-weighted interpolation across learned prototypes. 

\item \textbf{BLR (No Changepoint):} An ablation of our method that uses Bayesian Linear Regression (BLR) to update decoder weights online, while keeping the encoder fixed. The decoder is adapted using a streaming least-squares update with uncertainty propagation, but without any mechanism for changepoint detection. This variant isolates the contribution of changepoint-aware adaptation and allows us to assess how well BLR alone handles gradual model drift.
\end{enumerate}

\subsubsection{Ablation}
\label{app:ablation}
We evaluate the effect of beam size $K$ on prediction performance and computational efficiency of our changepoint-aware online adaptation framework. Larger beam sizes improve the model's capacity to track diverse hypotheses over latent shifts but increase inference cost. The aim of this ablation is to investigate this tradeoff.

We investigate how the number of tracked hypotheses (beam size \( K \)) affects both model accuracy and inference latency. Table~\ref{tab:ablation} reports results on the real-world quadrotor task under a 175g payload and forward velocity of \(1.5\)~m/s. We report:
\textbf{(A) Tracking RMSE} (in meters): lower is better, reflecting accuracy of the adaptive dynamics model. \textbf{(B) Time per Step} (in milliseconds): average model inference and update time per control step, excluding MPC solve time.

As \( K \) increases from 5 to 30, tracking RMSE consistently decreases, indicating improved adaptation accuracy due to better changepoint coverage. However, the gains diminish after \( K=15 \), with RMSE saturating near 0.118~m. This saturation is expected since additional hypotheses become redundant once the beam is large enough capture plausible changepoint trajectories. The predictive distribution of the model, computed via beam marginalization, stabilizes, and further increases to $K$ yield diminishing returns. 

Additionally, we observe that the inference latency remains nearly constant (around 7.6~ms) across all beam sizes with only marginal variation. This is due to the fact that all operations in our implementation are compiled by JAX/XLA into a single fused kernel, fully vectorized over the beam axis $K$. Beam-dependent computations such as marginal likelihood evaluation, Bayesian updates, and top-$K$ selection, are executed in parallel across SIMD lanes (on CPU) or streaming multiprocessors (on GPU). Consequently, the per-step latency is dominated by fixed kernel-launch and memory-access overheads, while the marginal cost of additional beams is negligible. Increasing $K$ merely utilizes idle compute lanes up to the device's hardware occupancy limit. This enables larger hypothesis sets ``for free'' in the cases considered in this work, with no measurable degradation in control frequency. 

\begin{table}[h]
\footnotesize
\renewcommand{\arraystretch}{1.3}
\caption{\small RMSE and Wall-time across Beam sizes.}
\centering
\scalebox{0.9}{
\begin{tabular}{lcc}
\toprule
\textbf{Beam Size} $K$ & \textbf{Time per step (ms)} & \textbf{RMSE (m)} ↓ \\
\midrule
5  & 7.43 & 0.128 \\
10 & 7.61 & 0.123 \\
15 & 7.66 & 0.120 \\
20 & 7.67 & 0.118 \\
30 & 7.69 & 0.118 \\
\bottomrule
\end{tabular}
}
\label{tab:ablation}
\end{table}
\vspace{-22pt}

\bibliographystyle{IEEEtran}
\bibliography{root}

@article{mayne2011tube,
  title={Tube-based robust nonlinear model predictive control},
  author={Mayne, David Q and Kerrigan, Erric C and Van Wyk, EJ and Falugi, Paola},
  journal={International journal of robust and nonlinear control},
  volume={21},
  number={11},
  pages={1341--1353},
  year={2011},
  publisher={Wiley Online Library}
}

@article{mesbah2016stochastic,
  title={Stochastic model predictive control: An overview and perspectives for future research},
  author={Mesbah, Ali},
  journal={IEEE Control Systems Magazine},
  volume={36},
  number={6},
  pages={30--44},
  year={2016},
  publisher={IEEE}
}

@article{torrente2021data,
  title={Data-driven mpc for quadrotors},
  author={Torrente, Guillem and Kaufmann, Elia and F{\"o}hn, Philipp and Scaramuzza, Davide},
  journal={IEEE Robotics and Automation Letters},
  volume={6},
  number={2},
  pages={3769--3776},
  year={2021},
  publisher={IEEE}
}

@article{chee2022knode,
  title={Knode-mpc: A knowledge-based data-driven predictive control framework for aerial robots},
  author={Chee, Kong Yao and Jiahao, Tom Z and Hsieh, M Ani},
  journal={IEEE Robotics and Automation Letters},
  volume={7},
  number={2},
  pages={2819--2826},
  year={2022},
  publisher={IEEE}
}

@inproceedings{chee2023enhancing,
  title={Enhancing sample efficiency and uncertainty compensation in learning-based model predictive control for aerial robots},
  author={Chee, Kong Yao and Silva, Thales C and Hsieh, M Ani and Pappas, George J},
  booktitle={2023 IEEE/RSJ International Conference on Intelligent Robots and Systems (IROS)},
  pages={9435--9441},
  year={2023},
  organization={IEEE}
}

@article{saviolo2022physics,
  title={Physics-inspired temporal learning of quadrotor dynamics for accurate model predictive trajectory tracking},
  author={Saviolo, Alessandro and Li, Guanrui and Loianno, Giuseppe},
  journal={IEEE Robotics and Automation Letters},
  volume={7},
  number={4},
  pages={10256--10263},
  year={2022},
  publisher={IEEE}
}

@inproceedings{gahlawat2020l1,
  title={L1-GP: L1 adaptive control with Bayesian learning},
  author={Gahlawat, Aditya and Zhao, Pan and Patterson, Andrew and Hovakimyan, Naira and Theodorou, Evangelos},
  booktitle={Learning for dynamics and control},
  pages={826--837},
  year={2020},
  organization={PMLR}
}

@article{o2022neural,
  title={Neural-fly enables rapid learning for agile flight in strong winds},
  author={O’Connell, Michael and Shi, Guanya and Shi, Xichen and Azizzadenesheli, Kamyar and Anandkumar, Anima and Yue, Yisong and Chung, Soon-Jo},
  journal={Science Robotics},
  volume={7},
  number={66},
  pages={eabm6597},
  year={2022},
  publisher={American Association for the Advancement of Science}
}

@inproceedings{jiahao2023online,
  title={Online dynamics learning for predictive control with an application to aerial robots},
  author={Jiahao, Tom Z and Chee, Kong Yao and Hsieh, M Ani},
  booktitle={Conference on Robot Learning},
  pages={2251--2261},
  year={2023},
  organization={PMLR}
}

@inproceedings{richards2021adaptive,
  title={Adaptive-Control-Oriented Meta-Learning for Nonlinear Systems},
  author={Richards, SM and Azizan, N and Slotine, J-JE and Pavone, M},
  booktitle={Robotics science and systems},
  year={2021}
}

@inproceedings{gu2024proto,
  title={Proto-MPC: An encoder-prototype-decoder approach for quadrotor control in challenging winds},
  author={Gu, Yuliang and Cheng, Sheng and Hovakimyan, Naira},
  booktitle={6th Annual Learning for Dynamics \& Control Conference},
  pages={1765--1776},
  year={2024},
  organization={PMLR}
}

@ARTICLE{saviolo2023active,
  author={Saviolo, Alessandro and Frey, Jonathan and Rathod, Abhishek and Diehl, Moritz and Loianno, Giuseppe},
  journal={IEEE Transactions on Robotics}, 
  title={Active Learning of Discrete-Time Dynamics for Uncertainty-Aware Model Predictive Control}, 
  year={2024},
  volume={40},
  number={},
  pages={1273-1291},
  doi={10.1109/TRO.2023.3339543}}

@inproceedings{joshi2019deep,
  title={Deep model reference adaptive control},
  author={Joshi, Girish and Chowdhary, Girish},
  booktitle={2019 IEEE 58th Conference on Decision and Control (CDC)},
  pages={4601--4608},
  year={2019},
  organization={IEEE}
}

@inproceedings{joshi2021asynchronous,
  title={Asynchronous deep model reference adaptive control},
  author={Joshi, Girish and Virdi, Jasvir and Chowdhary, Girish},
  booktitle={Conference on robot learning},
  pages={984--1000},
  year={2021},
  organization={PMLR}
}

@inproceedings{shi2019neural,
  title={Neural lander: Stable drone landing control using learned dynamics},
  author={Shi, Guanya and Shi, Xichen and O’Connell, Michael and Yu, Rose and Azizzadenesheli, Kamyar and Anandkumar, Animashree and Yue, Yisong and Chung, Soon-Jo},
  booktitle={2019 international conference on robotics and automation (icra)},
  pages={9784--9790},
  year={2019},
  organization={IEEE}
}

@inproceedings{djeumou2023learn,
  title={How to Learn and Generalize From Three Minutes of Data: Physics-Constrained and Uncertainty-Aware Neural Stochastic Differential Equations},
  author={Djeumou, Franck and Neary, Cyrus and Topcu, Ufuk},
  booktitle={Conference on Robot Learning},
  pages={577--601},
  year={2023},
  organization={PMLR}
}

@inproceedings{duong2021hamiltonian,
  title={Hamiltonian-based Neural ODE Networks on the SE (3) Manifold For Dynamics Learning and Control},
  author={Duong, Thai and Atanasov, Nikolay},
  booktitle={Robotics: Science and Systems (RSS)},
  year={2021}
}

@article{salzmann2023real,
  title={Real-time neural mpc: Deep learning model predictive control for quadrotors and agile robotic platforms},
  author={Salzmann, Tim and Kaufmann, Elia and Arrizabalaga, Jon and Pavone, Marco and Scaramuzza, Davide and Ryll, Markus},
  journal={IEEE Robotics and Automation Letters},
  volume={8},
  number={4},
  pages={2397--2404},
  year={2023},
  publisher={IEEE}
}

@article{fearnhead2007line,
  title={On-line inference for multiple changepoint problems},
  author={Fearnhead, Paul and Liu, Zhen},
  journal={Journal of the Royal Statistical Society Series B: Statistical Methodology},
  volume={69},
  number={4},
  pages={589--605},
  year={2007},
  publisher={Oxford University Press}
}

@article{adams2007bayesian,
  title={Bayesian online changepoint detection},
  author={Adams, Ryan Prescott and MacKay, David JC},
  journal={arXiv preprint arXiv:0710.3742},
  year={2007}
}

@article{ibrahim2000power,
  title={Power prior distributions for regression models},
  author={Ibrahim, Joseph G and Chen, Ming-Hui},
  journal={Statistical Science},
  pages={46--60},
  year={2000},
  publisher={JSTOR}
}

@article{ghosal2017fundamentals,
  title={Fundamentals of Nonparametric Bayesian Inference},
  author={Ghosal, Subhashis and van der Vaart, Aad},
  journal={Cambridge Books},
  year={2017},
  publisher={Cambridge University Press}
}

@book{murphy2023probabilistic,
  title={Probabilistic machine learning: Advanced topics},
  author={Murphy, Kevin P},
  year={2023}
}

@book{cesa2006prediction,
  title={Prediction, learning, and games},
  author={Cesa-Bianchi, Nicolo and Lugosi, G{\'a}bor},
  year={2006},
  publisher={Cambridge university press}
}

@book{matrixAnalysis,
author = {Horn, Roger A. and Johnson, Charles R.},
title = {Matrix analysis / Roger A. Horn, Charles R. Johnson.},
year = {2013},
address = {Cambridge},
booktitle = {Matrix analysis},
edition = {2nd ed.},
isbn = {9780521839402},
language = {eng},
publisher = {Cambridge University Press},
}

@article{optimistix,
    title={Optimistix: modular optimisation in JAX and Equinox},
    author={Jason Rader and Terry Lyons and Patrick Kidger},
    journal={arXiv:2402.09983},
    year={2024},
}

@software{jax2018github,
  author = {James Bradbury and Roy Frostig and Peter Hawkins and Matthew James Johnson and Chris Leary and Dougal Maclaurin and George Necula and Adam Paszke and Jake Vander{P}las and Skye Wanderman-{M}ilne and Qiao Zhang},
  title = {{JAX}: composable transformations of {P}ython+{N}um{P}y programs},
  url = {http://github.com/jax-ml/jax},
  version = {0.3.13},
  year = {2018},
}

@article{brunton2016discovering,
  title={Discovering governing equations from data by sparse identification of nonlinear dynamical systems},
  author={Brunton, Steven L and Proctor, Joshua L and Kutz, J Nathan},
  journal={Proceedings of the national academy of sciences},
  volume={113},
  number={15},
  pages={3932--3937},
  year={2016},
  publisher={National Academy of Sciences}
}

@article{kaiser2018sparse,
  title={Sparse identification of nonlinear dynamics for model predictive control in the low-data limit},
  author={Kaiser, Eurika and Kutz, J Nathan and Brunton, Steven L},
  journal={Proceedings of the Royal Society A},
  volume={474},
  number={2219},
  pages={20180335},
  year={2018},
  publisher={The Royal Society Publishing}
}

@inproceedings{kurle2019continual,
  title={Continual learning with bayesian neural networks for non-stationary data},
  author={Kurle, Richard and Cseke, Botond and Klushyn, Alexej and Van Der Smagt, Patrick and G{\"u}nnemann, Stephan},
  booktitle={International Conference on Learning Representations},
  year={2019}
}

@article{ramos2019bayessim,
  title={BayesSim: Adaptive Domain Randomization Via Probabilistic Inference for Robotics Simulators},
  author={Ramos, Fabio and Possas, Rafael and Fox, Dieter},
  journal={Robotics: Science and Systems XV},
  year={2019},
  publisher={Robotics: Science and Systems Foundation}
}

@inproceedings{wu2023daydreamer,
  title={Daydreamer: World models for physical robot learning},
  author={Wu, Philipp and Escontrela, Alejandro and Hafner, Danijar and Abbeel, Pieter and Goldberg, Ken},
  booktitle={Conference on robot learning},
  pages={2226--2240},
  year={2023},
  organization={PMLR}
}

@inproceedings{hafner2019learning,
  title={Learning latent dynamics for planning from pixels},
  author={Hafner, Danijar and Lillicrap, Timothy and Fischer, Ian and Villegas, Ruben and Ha, David and Lee, Honglak and Davidson, James},
  booktitle={International conference on machine learning},
  pages={2555--2565},
  year={2019},
  organization={PMLR}
}

@article{mania2022active,
  title={Active learning for nonlinear system identification with guarantees},
  author={Mania, Horia and Jordan, Michael I and Recht, Benjamin},
  journal={Journal of Machine Learning Research},
  volume={23},
  number={32},
  pages={1--30},
  year={2022}
}

@article{li2021detecting,
  title={Detecting and adapting to irregular distribution shifts in bayesian online learning},
  author={Li, Aodong and Boyd, Alex and Smyth, Padhraic and Mandt, Stephan},
  journal={Advances in neural information processing systems},
  volume={34},
  pages={6816--6828},
  year={2021}
}

@article{knoblauch2018doubly,
  title={Doubly Robust Bayesian Inference for Non-Stationary Streaming Data with $\beta$-Divergences},
  author={Knoblauch, Jeremias and Jewson, Jack E and Damoulas, Theodoros},
  journal={Advances in Neural Information Processing Systems},
  volume={31},
  year={2018}
}

@inproceedings{saatcci2010gaussian,
  title={Gaussian process change point models},
  author={Saat{\c{c}}i, Yunus and Turner, Ryan D and Rasmussen, Carl E},
  booktitle={Proceedings of the 27th International Conference on Machine Learning (ICML-10)},
  pages={927--934},
  year={2010}
}

@inproceedings{nguyen2018variational,
  title={Variational Continual Learning},
  author={Nguyen, Cuong V and Li, Yingzhen and Bui, Thang D and Turner, Richard E},
  booktitle={International Conference on Learning Representations},
  year={2018}
}

@article{feng2022factored,
  title={Factored adaptation for non-stationary reinforcement learning},
  author={Feng, Fan and Huang, Biwei and Zhang, Kun and Magliacane, Sara},
  journal={Advances in Neural Information Processing Systems},
  volume={35},
  pages={31957--31971},
  year={2022}
}

@article{saviolo2023learning,
  title={Learning quadrotor dynamics for precise, safe, and agile flight control},
  author={Saviolo, Alessandro and Loianno, Giuseppe},
  journal={Annual Reviews in Control},
  volume={55},
  pages={45--60},
  year={2023},
  publisher={Elsevier}
}

@article{lambert2022investigating,
  title={Investigating compounding prediction errors in learned dynamics models},
  author={Lambert, Nathan and Pister, Kristofer and Calandra, Roberto},
  journal={arXiv preprint arXiv:2203.09637},
  year={2022}
}

@article{lew2022safe,
  title={Safe active dynamics learning and control: A sequential exploration--exploitation framework},
  author={Lew, Thomas and Sharma, Apoorva and Harrison, James and Bylard, Andrew and Pavone, Marco},
  journal={IEEE Transactions on Robotics},
  volume={38},
  number={5},
  pages={2888--2907},
  year={2022},
  publisher={IEEE}
}

@inproceedings{harvey1995limited,
  title={Limited discrepancy search},
  author={Harvey, William D and Ginsberg, Matthew L},
  booktitle={Proceedings of the 14th international joint conference on Artificial intelligence-Volume 1},
  pages={607--613},
  year={1995}
}

@ARTICLE{wu2023l1,
  author={Wu, Zhuohuan and Cheng, Sheng and Zhao, Pan and Gahlawat, Aditya and Ackerman, Kasey A. and Lakshmanan, Arun and Yang, Chengyu and Yu, Jiahao and Hovakimyan, Naira},
  journal={IEEE Transactions on Control Systems Technology}, 
  title={L1Quad: L1 Adaptive Augmentation of Geometric Control for Agile Quadrotors With Performance Guarantees}, 
  year={2025},
  volume={33},
  number={2},
  pages={597-612},
  doi={10.1109/TCST.2024.3521182}}

@inproceedings{wan2000unscented,
  title={The unscented Kalman filter for nonlinear estimation},
  author={Wan, Eric A and Van Der Merwe, Rudolph},
  booktitle={Proceedings of the IEEE 2000 adaptive systems for signal processing, communications, and control symposium (Cat. No. 00EX373)},
  pages={153--158},
  year={2000},
  organization={Ieee}
}

@ARTICLE{hersch2008,
  author={Hersch, Micha and Guenter, Florent and Calinon, Sylvain and Billard, Aude},
  journal={IEEE Transactions on Robotics}, 
  title={Dynamical System Modulation for Robot Learning via Kinesthetic Demonstrations}, 
  year={2008},
  volume={24},
  number={6},
  pages={1463-1467},
  doi={10.1109/TRO.2008.2006703}}

@article{zhou2025simultaneous,
  title={Simultaneous system identification and model predictive control with no dynamic regret},
  author={Zhou, Hongyu and Tzoumas, Vasileios},
  journal={IEEE Transactions on Robotics},
  year={2025},
  publisher={IEEE}
}

@ARTICLE{wei2025mlmpcc,
  author={Wei, Mingxin and Zheng, Lanxiang and Wu, Ying and Mei, Ruidong and Cheng, Hui},
  journal={IEEE Transactions on Robotics}, 
  title={Meta-Learning Enhanced Model Predictive Contouring Control for Agile and Precise Quadrotor Flight}, 
  year={2025},
  volume={41},
  number={},
  pages={3590-3608},
  doi={10.1109/TRO.2025.3567491}}

@article{jia2023evolver,
  title={Evolver: Online learning and prediction of disturbances for robot control},
  author={Jia, Jindou and Zhang, Wenyu and Guo, Kexin and Wang, Jianliang and Yu, Xiang and Shi, Yang and Guo, Lei},
  journal={IEEE Transactions on Robotics},
  volume={40},
  pages={382--402},
  year={2023},
  publisher={IEEE}
}

@article{mckinnon2021meta,
  title={Meta learning with paired forward and inverse models for efficient receding horizon control},
  author={McKinnon, Christopher D and Schoellig, Angela P},
  journal={IEEE Robotics and Automation Letters},
  volume={6},
  number={2},
  pages={3240--3247},
  year={2021},
  publisher={IEEE}
}

@article{belkhale2021model,
  title={Model-based meta-reinforcement learning for flight with suspended payloads},
  author={Belkhale, Suneel and Li, Rachel and Kahn, Gregory and McAllister, Rowan and Calandra, Roberto and Levine, Sergey},
  journal={IEEE Robotics and Automation Letters},
  volume={6},
  number={2},
  pages={1471--1478},
  year={2021},
  publisher={IEEE}
}

@article{park2020gaussian,
  title={Gaussian process online learning with a sparse data stream},
  author={Park, Jehyun and Choi, Jongeun},
  journal={IEEE Robotics and Automation Letters},
  volume={5},
  number={4},
  pages={5977--5984},
  year={2020},
  publisher={IEEE}
}

@inproceedings{chakrabarty2024physics,
  title={Physics-constrained meta-learning for online adaptation and estimation in positioning applications},
  author={Chakrabarty, Ankush and Deshpande, Vedang and Wichern, Gordon and Berntorp, Karl},
  booktitle={2024 IEEE 63rd Conference on Decision and Control (CDC)},
  pages={1561--1566},
  year={2024},
  organization={IEEE}
}

@INPROCEEDINGS{1657243,
  author={Chengyu Cao and Hovakimyan, N.},
  booktitle={2006 American Control Conference}, 
  title={Design and Analysis of a Novel L1 Adaptive Controller, Part I: Control Signal and Asymptotic Stability}, 
  year={2006},
  volume={},
  number={},
  pages={3397-3402},
  doi={10.1109/ACC.2006.1657243}}

@book{hovakimyan2010,
  title={L1 adaptive control theory: Guaranteed robustness with fast adaptation},
  author={Hovakimyan, Naira and Cao, Chengyu},
  year={2010},
  publisher={SIAM}
}

@ARTICLE{8909368,
  author={Hewing, Lukas and Kabzan, Juraj and Zeilinger, Melanie N.},
  journal={IEEE Transactions on Control Systems Technology}, 
  title={Cautious Model Predictive Control Using Gaussian Process Regression}, 
  year={2020},
  volume={28},
  number={6},
  pages={2736-2743},
  doi={10.1109/TCST.2019.2949757}}

@ARTICLE{10214438,
  author={Xu, Shaohang and Zhu, Lijun and Zhang, Hai-Tao and Ho, Chin Pang},
  journal={IEEE Transactions on Robotics}, 
  title={Robust Convex Model Predictive Control for Quadruped Locomotion Under Uncertainties}, 
  year={2023},
  volume={39},
  number={6},
  pages={4837-4854},
  doi={10.1109/TRO.2023.3299527}}

@ARTICLE{9719129,
  author={Song, Yunlong and Scaramuzza, Davide},
  journal={IEEE Transactions on Robotics}, 
  title={Policy Search for Model Predictive Control With Application to Agile Drone Flight}, 
  year={2022},
  volume={38},
  number={4},
  pages={2114-2130},
  doi={10.1109/TRO.2022.3141602}}

@ARTICLE{10049101,
  author={Salzmann, Tim and Kaufmann, Elia and Arrizabalaga, Jon and Pavone, Marco and Scaramuzza, Davide and Ryll, Markus},
  journal={IEEE Robotics and Automation Letters}, 
  title={Real-Time Neural MPC: Deep Learning Model Predictive Control for Quadrotors and Agile Robotic Platforms}, 
  year={2023},
  volume={8},
  number={4},
  pages={2397-2404},
  doi={10.1109/LRA.2023.3246839}}

@INPROCEEDINGS{10611492,
  author={Sacks, Jacob and Rana, Rwik and Huang, Kevin and Spitzer, Alex and Shi, Guanya and Boots, Byron},
  booktitle={2024 IEEE International Conference on Robotics and Automation (ICRA)}, 
  title={Deep Model Predictive Optimization}, 
  year={2024},
  volume={},
  number={},
  pages={16945-16953},
  doi={10.1109/ICRA57147.2024.10611492}
}

@article{jastrzkebski2017three,
  title={Three factors influencing minima in sgd},
  author={Jastrzebski, Stanis{\l}aw and Kenton, Zachary and Arpit, Devansh and Ballas, Nicolas and Fischer, Asja and Bengio, Yoshua and Storkey, Amos},
  journal={arXiv preprint arXiv:1711.04623},
  year={2017}
}

@inproceedings{smith2017cyclical,
  title={Cyclical learning rates for training neural networks},
  author={Smith, Leslie N},
  booktitle={2017 IEEE winter conference on applications of computer vision (WACV)},
  pages={464--472},
  year={2017},
  organization={IEEE}
}

@inproceedings{song2020rapidly,
  title={Rapidly adaptable legged robots via evolutionary meta-learning},
  author={Song, Xingyou and Yang, Yuxiang and Choromanski, Krzysztof and Caluwaerts, Ken and Gao, Wenbo and Finn, Chelsea and Tan, Jie},
  booktitle={2020 IEEE/RSJ International Conference on Intelligent Robots and Systems (IROS)},
  pages={3769--3776},
  year={2020},
  organization={IEEE}
}

@inproceedings{kaushik2020fast,
  title={Fast online adaptation in robotics through meta-learning embeddings of simulated priors},
  author={Kaushik, Rituraj and Anne, Timoth{\'e}e and Mouret, Jean-Baptiste},
  booktitle={2020 IEEE/RSJ International Conference on Intelligent Robots and Systems (IROS)},
  pages={5269--5276},
  year={2020},
  organization={IEEE}
}

@inproceedings{das2025dronediffusion,
  title={Dronediffusion: Robust quadrotor dynamics learning with diffusion models},
  author={Das, Avirup and Yadav, Rishabh Dev and Sun, Sihao and Sun, Mingfei and Kaski, Samuel and Pan, Wei},
  booktitle={2025 IEEE International Conference on Robotics and Automation (ICRA)},
  pages={1604--1610},
  year={2025},
  organization={IEEE}
}

@inproceedings{cao2024computation,
  author    = {Wenhan Cao and Alexandre Capone and Rishabh Yadav and Sandra Hirche and Wei Pan},
  title     = {Computation-Aware Learning for Stable Control with Gaussian Process},
  booktitle = {Proceedings of Robotics: Science and Systems (RSS)},
  year      = {2024}
}
\end{document}